\pdfoutput=1

\documentclass[11pt]{article}

\usepackage[preprint]{acl}
\usepackage{float}
\usepackage{graphicx}    
\usepackage{subcaption}  
\usepackage{hyperref} 
\usepackage{xcolor}
\usepackage{multirow}
\usepackage{booktabs}
\usepackage{times}
\usepackage{latexsym}
\usepackage{tcolorbox}
\usepackage[T1]{fontenc}

\usepackage[utf8]{inputenc}

\usepackage{microtype}

\usepackage{inconsolata}
\usepackage{enumitem}
\usepackage{times}
\usepackage{latexsym}

\usepackage[T1]{fontenc}

\usepackage[utf8]{inputenc}

\usepackage{microtype}

\usepackage{inconsolata}

\usepackage{graphicx}

\title{The Power of Personality: A Human Simulation Perspective to Investigate Large Language Model Agents}

\author{Yifan Duan\footnotemark[1], Yihong Tang\footnotemark[1], Xuefeng Bai, \\ {\bf Kehai Chen\footnotemark[2], Juntao Li, Min Zhang} \\
        Institute of Computing and Intelligence, Harbin Institute of Technology, Shenzhen, China \\
        \texttt{\{ziye\_0926@outlook.com,  chenkehai@hit.edu.cn\}}
}

\begin{document}
\maketitle
\begingroup\def\thefootnote{*}\footnotetext{Equal contribution.}\endgroup
\begingroup\def\thefootnote{\dag}\footnotetext{Corresponding author.}\endgroup
\begin{abstract}
Large language models (LLMs) excel in both closed tasks (including problem-solving, and code generation) and open tasks (including creative writing), yet existing explanations for their capabilities lack connections to real-world human intelligence. 
To fill this gap, this paper systematically investigates LLM intelligence through the lens of ``human simulation'', addressing three core questions: (1) \textit{How do personality traits affect problem-solving in closed tasks?} (2) \textit{How do traits shape creativity in open tasks?} (3) \textit{How does single-agent performance influence multi-agent collaboration?} 
By assigning Big Five personality traits to LLM agents and evaluating their performance in single- and multi-agent settings, we reveal that specific traits significantly influence reasoning accuracy (closed tasks) and creative output (open tasks). 
Furthermore,  multi-agent systems exhibit collective intelligence distinct from individual capabilities, driven by distinguishing combinations of personalities.
\end{abstract}

\section{Introduction}


Large Language Models (LLMs) have demonstrated exceptional performance and broad application potential in closed tasks such as knowledge question answering, and mathematical reasoning, as well as open-ended tasks like article writing, and poetry creation~\cite{lauriola-etal-2025-analyzing,shao-etal-2024-assisting,chen-etal-2024-evaluating-diversity}. Existing research has explored explanations for LLMs' wide-ranging capabilities from multiple perspectives, including context learning mechanisms, emergent model scaling properties, and theories of compressed intelligence~\cite{Schaeffer2023AreEA,Liskavets2024PromptCW,Akyrek2022WhatLA}. These studies primarily focus on the intrinsic properties and training dynamics of the models.

While existing explanations help understand how LLMs function, connecting their capabilities to human intelligence and real-world behavior remains a challenge. To explore this connection, researchers are exploring LLMs by simulating human behaviors and contexts. 
For example, models can enhance performance by simulating human thought processes~\cite{Wei2022ChainOT} and collaborative discussions~\cite{Du2023ImprovingFA,Chan2024ChatEvalTB}. 
Assigning specific roles through role-playing has also been found to improve model performance in both closed-ended and open-ended tasks~\cite{kong2024betterzeroshotreasoningroleplay,Lu2024LLMDE}.

However, while existing works have preliminarily explored the impact of simulating human behavior and roles, the potential effect of human psychological traits, a core element of human on model capabilities has not yet been sufficiently studied. To fill this research gap, this study systematically examines the performance patterns of the model when simulating different personality traits, based on the Big Five personality traits theory in psychology~\cite{Goldberg1990AnA}.This study aims to preliminarily address the following key questions:

\begin{itemize}[itemsep=2pt,topsep=0pt,parsep=0pt]
\item \textit{In closed tasks, how does simulating different personality traits affect the capability performance of agents?}
\item \textit{In open-ended tasks, how does simulating different personality traits affect the creativity of agents?}
\item \textit{In multi-agent collaborative environments, does simulating different personality traits influence team collaboration performance?}
\end{itemize}

To answer these questions, this study examines the impact of simulated personality traits on LLMs at both single-agent and multi-agent levels. First, we assign different Big Five personality traits to agents and validate the accuracy and degree of their simulated personalities through psychological scales. Subsequently, we evaluate the performance differences of these agents with different personalities in both closed and open tasks, exploring how personality traits affect problem-solving ability and creativity, and comparing findings with real psychological research conclusions. Finally, we construct agent teams with different personality trait combinations and analyze the influence of personality traits on collaborative effectiveness through a multi-agent collaboration framework.

Finally, this paper preliminarily answers the three questions: (1) In closed tasks, different personality traits cause performance differences in agents, and the influence of personality traits on agents shows similarities with some psychological findings. (2) In open-ended tasks, different personality traits influence agent creativity, and the relationship between personality traits and creativity in models shows certain similarities with reality. (3) Personality traits influence multi-agent collaborative performance, with impacts similar to real-world situations, and specific personality trait combinations can further Influence multi-agent collaborative performance. These findings elaborate on the influence and exploratory potential of human simulation on model capabilities from the perspective of human psychology, providing reference for further exploring and enhancing model capabilities from the perspective of human simulation.

\section{Related Work}

\paragraph{Human Simulation} Researchers have utilized agents to simulate social environments, enabling the reproduction of real-world interactions and behavioral patterns in virtual scenarios with low cost and high efficiency~\cite{stanfordTown,Chen2023AgentVerseFM,liu2024lmagentlargescalemultimodalagents,li-etal-2024-econagent}. 
In this process, agents not only possess autonomous decision-making capabilities but also exhibit complex personalized traits, such as cognitive abilities, emotional responses, motivations, and values~\cite{Li2024EvolvingAI,pmlr-v235-choi24e,he2024afsppagentframeworkshaping}. These personalized characteristics influence the behavioral choices and decision-making processes of agents. 
By simulating human behaviors and roles, existing studies have further influenced and enhanced the performance of agents~\cite{kong2024betterzeroshotreasoningroleplay,Wei2022ChainOT,Chen2024HoLLMwoodUT}. Our work explores the impact of simulating personality traits on model capabilities from the perspective of human psychological simulation.

\paragraph{Personality in LLMs} Recent studies indicate that large language models (LLMs) can effectively adopt specific personality traits~\citep{frisch-giulianelli-2024-llm,jiang-etal-2024-personallm}. For instance, ~\citet{wang2024incharacterevaluatingpersonalityfidelity} and \citet{vijjini2024exploringsafetyutilitytradeoffspersonalized} observed that LLMs prompted with different personalities achieve scores on self-reported Big Five personality questionnaires that closely align with their assigned traits. This underscores the plasticity of LLMs in expressing and simulating personality traits. However, other research suggests that while LLMs can be somewhat influenced by trait specifications in prompts, they often maintain an intrinsic "personality type"~\cite{lacava2025openmodelsclosedminds}. \citet{Caron2023ManipulatingTP} investigated how LLM outputs change when prompted with different personality traits, finding that the models can indeed reflect the implied personalities in their responses. Building on this work, ~\citet{Zhu2024TraitsPromptDP} further explored how different personalities affect LLM performance across a range of tasks. Our research focuses on human psychological simulation, enabling LLMs to mimic diverse personality traits. We investigate how simulating these traits impacts model performance on various tasks, and analyze the parallels between these observed effects and actual human performance.

\begin{figure*}[htbp]
    \centering
    \includegraphics[width=1\textwidth]{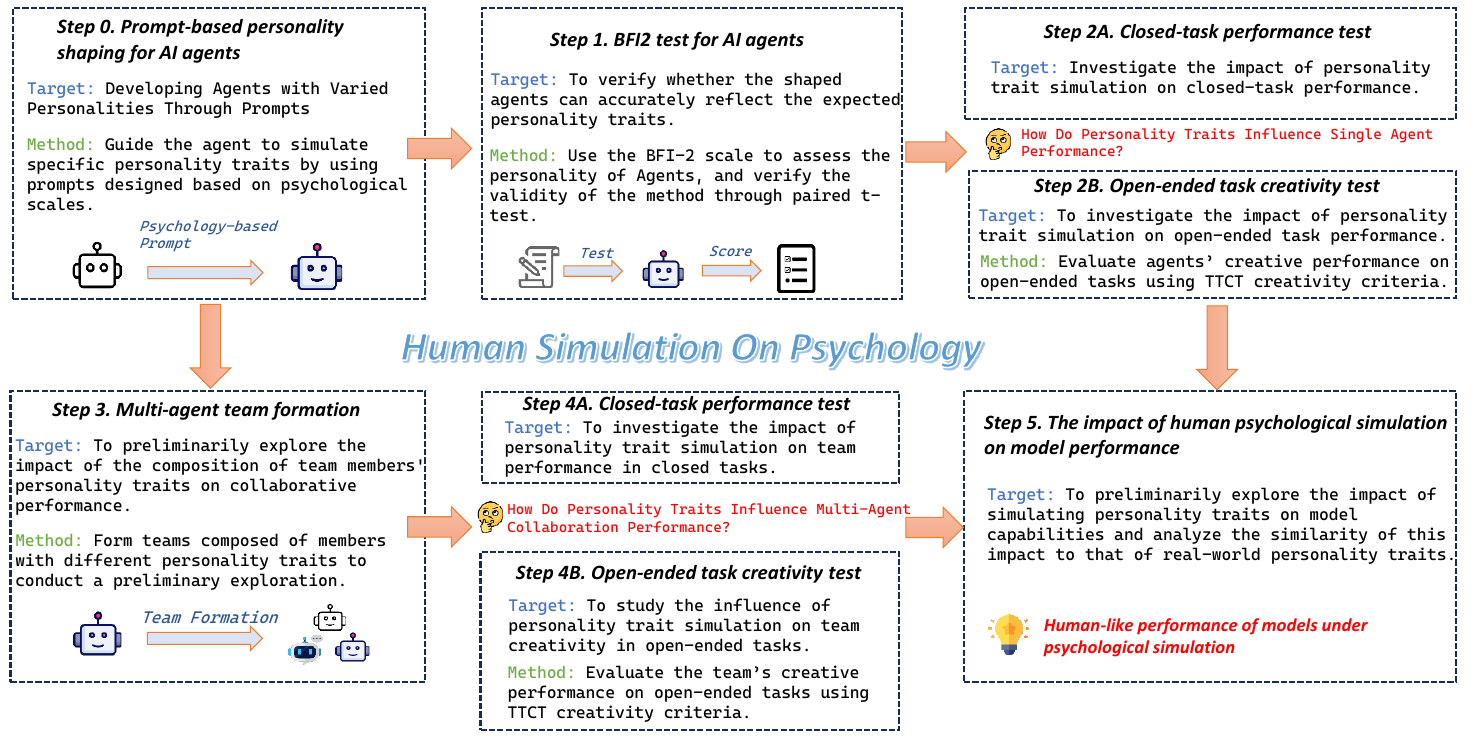}
    \caption{Illustration of the overall framework: First, we set prompts based on psychological scales to guide the agent in simulating different Big Five personality traits. Then, we use the BFI-2 scale to test whether the agent can accurately exhibit the designated personality traits. Next, we test agents with different personality traits in both closed and open tasks to explore the impact of simulating different personality traits on agent performance. In addition, we form teams composed of agents with different personality traits and test these teams in both closed and open tasks to study the effect of simulated personality traits on team effectiveness.}
    \label{intro}
\end{figure*}

\section{Method}
This section introduces the methods for enabling LLM agents to simulate the Big Five personality traits, as well as the approaches for testing the accuracy of their simulated personalities and configuring multi-agent collaboration. We provide a detailed explanation of how personalized prompts are used to embed different personality traits into the agents, and describe the experimental setup for evaluating agent performance in both closed and open tasks.  
All methodological details are provided in Appendix~\ref{config}.
\subsection{Configuration of Big Five Personality Traits for Agents}
\label{Configuration}

We use the scale items proposed by ~\citet{DeYoung2007BetweenFA} to set personalized instructions and embed them into the context to construct LLM agents with different Big Five personality traits. For example, using "Rarely get irritated" in the prompt will result in an agent with low Neuroticism. Detail prompt can be found in Appendix~\ref{app: prompts}.

For each LLM agent \(k\), its personality traits \(y_k\) can be represented as the following function:
\[
    y_k = f(X_{NEU}, X_{AGR}, X_{CON}, X_{EXT}, X_{OPE}).
    \label{eq:psi_definition}
\]

Where \(X \in \{-, +\}\) represents the polarity of each personality dimension (where \(-\) indicates negative traits and \(+\) indicates positive traits). The function \(f\) describes the agent's composite characteristics across the five dimensions: Neuroticism (NEU), Agreeableness (AGR), Conscientiousness (CON), Extraversion (EXT), and Openness (OPE). For example, \(y_k = f(-----)\) indicates that the LLM agent exhibits negative traits across all five dimensions.

Through the approach above, we examine combinations of high and low polarities across each personality dimension, thereby creating a total of \(2^5=32\) agents with distinct personality traits. 
\subsection{Personality Test}

We use the BFI-2 scale~\cite{Soto2017TheNB} to investigate whether LLMs can demonstrate and to what extent they exhibit the personality traits assigned through our methods. The BFI-2 is a widely recognized personality assessment tool in psychological research. It consists of 60 items that comprehensively measure an individual's personality across five major dimensions.  Further information can be found in Appendix~\ref{BFI-2}.

We administer the BFI-2 test to each LLM agent with different personality configurations by presenting standardized test questions. We then collect and score their responses to verify whether these artificial agents consistently display the specific personality traits specified in their prompts.

\subsection{Multi-agent Collaboration}
\textit{In a crowd, the individual is submerged.} To study how personality traits affect multi-agent collaboration, we use a basic LLM debate framework to explore agent behaviors and team performance in both closed and open tasks~\cite{Du2023ImprovingFA,Lu2024LLMDE}.
In our experiment, the agent team is formed based on different combinations of Big Five personality traits. The entire discussion process is divided into three stages: initial answer generation, group discussion, and final answer generation.

Agents with distinct personality traits initially generate diverse responses. Then agents share their answers and discuss together to reach a team consensus, with their varying personality traits shaping the dynamics of interaction and decision-making. After discussion, each agent refines its answer by incorporating insights from the group, and all final answers of agents are combined into a collective team answer.

\subsection{Evaluation}
To evaluate how the Big Five personality traits affect agent performance, we study both closed and open tasks. 

\paragraph{Evaluation of Closed Tasks}
Closed tasks refer to objective tasks with clear solutions and answers, such as answering factual questions. We select representative benchmarks to evaluate the performance of agents with different Big Five personality traits in closed tasks. MMLU ~\cite{Hendrycks2020MeasuringMM} is a large benchmark covering 57 subjects to assess a model's knowledge and reasoning abilities. MMLU-Pro ~\cite{Wang2024MMLUProAM}, an advanced version, includes more challenging tasks focusing on the depth of knowledge and problem-solving skills in professional domains. ARC ~\cite{allenai:arc} tests basic knowledge and advanced reasoning skills. SciQ ~\cite{SciQ} evaluates a model's understanding and reasoning of scientific concepts . Additionally, in multi-agent collaboration, we introduce GPQA~\cite{Rein2023GPQAAG} for evaluation to explore the impact of personality traits on collaboration when facing high-difficulty problems. Based on these test results, we systematically analyze the performance of models with different personality traits and study how personality influences agents' performance in closed tasks.

\paragraph{Evaluation of Open Tasks}
Open tasks are inherently subjective and lack standardized answers, making creativity a crucial aspect to evaluate. We assess this using the Torrance Tests of Creative Thinking (TTCT)~\cite{torrance1966}, a reliable framework evaluating creativity across four dimensions: Originality (idea novelty), Elaboration (detail), Fluency (quantity), and Flexibility (category variety). In Appendix~\ref{TTCT},we show more details of TTCT. Following ~\citet{Lu2024LLMDE}, our assessment incorporates common creativity benchmarks: AUT, INSTANCES, SIMILARITIES, and SCIENTIFIC. To operationalize the evaluation, we employ gpt-4o-mini as an automated assessor, guided by recent approaches~\cite{DiStefano2024AutomaticSO,Lu2024LLMDE}. We score each generated response along the four TTCT dimensions, providing a systematic measure of creative performance. Appendix~\ref{eva} contains the complete evaluation methodology.

\subsection{Experiments Setup}
We conduct our experiments using three large language models: Qwen2.5-14B-Instruct, Qwen2.5-32B-Instruct~\cite{qwen2.5}, and Llama3.1-8B-Instruct~\cite{meta2024llama3}, with the temperature set to 0 to ensure deterministic results. All models are deployed using vLLM~\cite{kwon2023efficient}, and inference is performed on NVIDIA L20 GPUs. We use gpt-4o-mini for creativity assessment through API calls, with the temperature set to 0.

\section{Personality's Influence on Single-Agent}

\subsection{BFI-2 Score}
We employ the BFI-2 personality scale to evaluate all the shaped agents. The detailed test scores for different agents across various models are fully presented in Appendix~\ref{Experiment}. To verify whether our method can significantly shape corresponding polar personality traits across different dimensions, we conduct paired t-tests, with results shown in Table~\ref{tab:ttest_results}.

\begin{table*}
\centering
\resizebox{\textwidth}{!}{
\begin{tabular}{lccccccccccccccc}
\hline
\multirow{2}{*}{Model} & \multicolumn{3}{c}{NEG} & \multicolumn{3}{c}{AGE} & \multicolumn{3}{c}{CON} & \multicolumn{3}{c}{EXT} & \multicolumn{3}{c}{OPE} \\
\cline{2-16}
& t & p & Cohen's d & t & p & Cohen's d & t & p &Cohen's d & t & p & Cohen's d & t & p & Cohen's d \\
\hline
Qwen-32B & 131.08 & <0.001 & 23.17 & 21.38 & <0.001 & 3.78 & 22.63 & <0.001 & 4.00 & 19.04 & <0.001 & 3.37 & 12.09 & <0.001 & 2.14 \\
Qwen-14B& 104.48 & <0.001 & 18.47 & 14.09 & <0.001 & 2.49 & 11.63 & <0.001 & 2.06 & 16.23 & <0.001 & 2.87 & 16.25 & <0.001 & 2.87 \\
Llama-8B& 19.07 & <0.001 & 3.37 & 8.40 & <0.001 & 1.49 & 9.89 & <0.001 & 1.75 & 20.74 & <0.001 & 3.67 & 9.93 & <0.001 & 1.75 \\
\bottomrule
\end{tabular}
}
\caption{Paired t-test Results for Big Five Personality Traits Across Different LLMs}
\label{tab:ttest_results}
\end{table*}

Experimental results show that for all five personality dimensions across all models, the p-values from the paired t-tests are significantly less than 0.001. This suggests that the method we employ is highly statistically significant in shaping the corresponding personality traits of the agents. Furthermore, Cohen's d, as a measure of effect size, exhibit large effect sizes under all conditions, further confirming the practical effectiveness of this shaping method.

In the comparison between models, we observe differences in performance. Qwen-32B generally exhibits the highest t-values and Cohen's d values across all dimensions, indicating the strongest shaping effect. The performance of Qwen-14B is also robust, with its effect sizes generally surpassing those of Llama-8B. In contrast, although Llama-8B achieve statistical significance and possess large effect sizes in most cases, its t-values and Cohen's d values are relatively lower. Notably, Llama-8B encounter challenges when shaping specific combinations of personality traits. For instance, when attempting to shape the personality represent as $y_k=f(-+-+-)$, its Conscientiousness score is actually 3.25, failing to meet the expected low score standard. This suggests that the effectiveness of our method in simulating personality traits may be related to the underlying capabilities of the model itself.

\subsection{Closed Tasks}

In Appendix~\ref{apd:resultofclose}, we present the detailed performance results of models simulating different Big Five personality traits on closed tasks. The data indicates that agents' performance on these tasks varies significantly depending on the personality traits. 

\begin{figure*}[ht]
    \centering
    \begin{subfigure}[b]{0.32\textwidth}
        \centering
        \includegraphics[width=\textwidth]{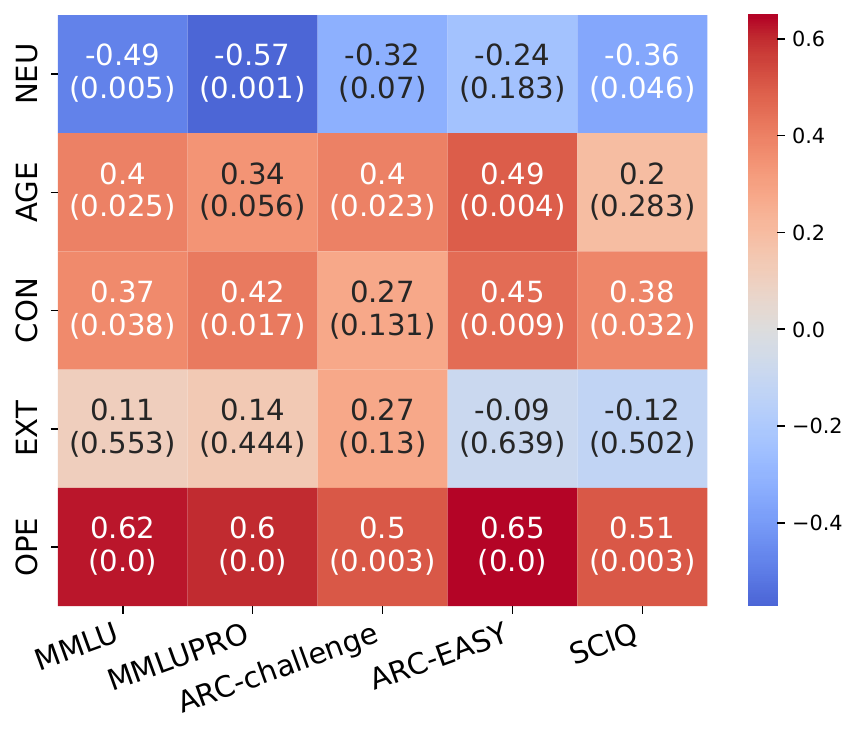}
        \caption{Qwen2.5-32B}
        \label{fig:subfig1}
    \end{subfigure}
    \hfill  
    \begin{subfigure}[b]{0.32\textwidth}
        \centering
        \includegraphics[width=\textwidth]{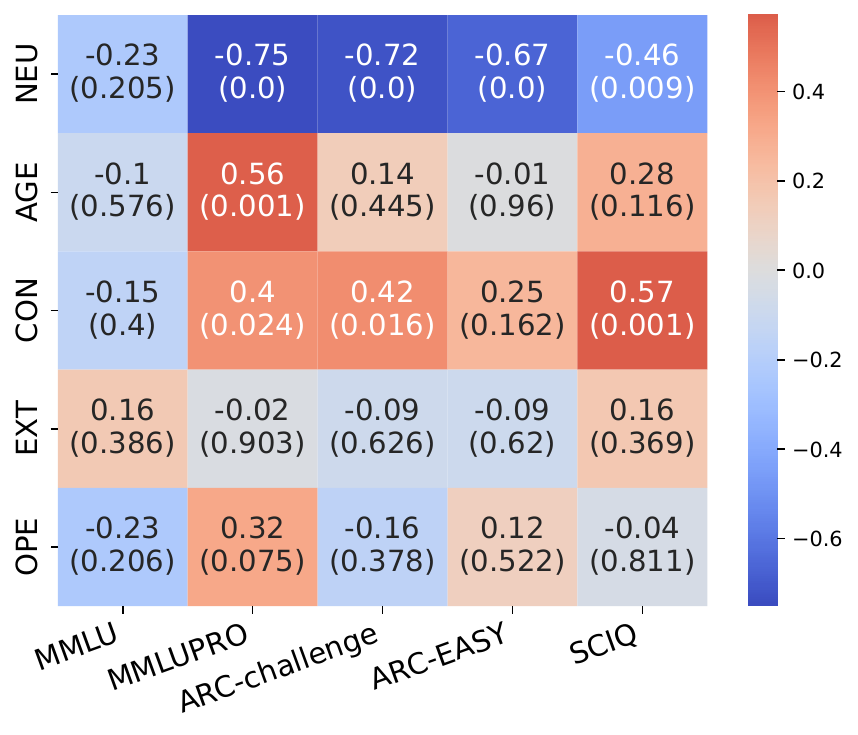}
        \caption{Qwen2.5-14B}
        \label{fig:subfig2}
    \end{subfigure}
    \hfill  
    \begin{subfigure}[b]{0.32\textwidth}
        \centering
        \includegraphics[width=\textwidth]{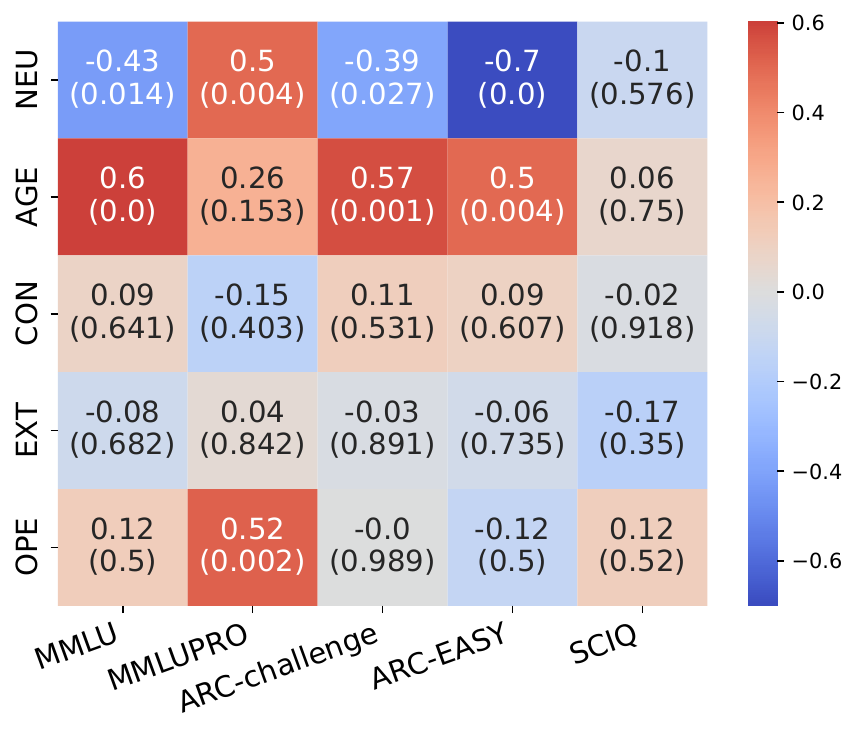}
        \caption{Llama3.1-8B}
        \label{fig:subfig3}
    \end{subfigure}
    \caption{Correlation Analysis between Personality Traits and Closed Tasks: Pearson Correlation Coefficients and Statistical Significance (p-values) between Five Personality Dimensions and the Accuracy of Five Closed Tasks.}
    \label{fig:three_images}
\end{figure*}

\paragraph{Personality and Accuracy}
Psychological research suggests that individual personality traits have certain effects on problem-solving abilities: high conscientiousness, high openness, and positive psychological states often predict stronger problem-solving capabilities, while high neuroticism is typically associated with weaker problem-solving efficacy~\cite{Babaei2018RelationshipBB,ElOthman2020PersonalityTE,Kipman2022PersonalityTA}. To verify whether this psychological phenomenon is also manifested in LLMs, we conduct correlation analyses on agents with different personality traits, calculating correlation coefficients between personality trait dimension scores and task accuracy rates to preliminarily explore the influence of personality factors on models' problem-solving abilities. The results of these correlation analyses are shown in Figure~\ref{fig:three_images}.

Consistent with human studies, Neuroticism scores generally exhibit a negative correlation with task accuracy across most tasks and all models, often reaching statistical significance. For example, Qwen-14B show significant negative correlations across all five tasks, and Llama-8B display significant negative correlations on MMLU and ARC. This suggests that simulating higher neuroticism might indeed inhibit cognitive performance in these models. However, there are notable exceptions, such as a significant positive correlation observed between NEU and accuracy for Llama3.1-8B specifically on MMLU-Pro.

For openness, its impact on LLMs vary significantly. Qwen-32B aligns with human findings, showing strong, significant positive correlations across all tasks. In contrast, Qwen-14B displays generally weak and non-significant correlations, sometimes slightly negative. Llama-8B only shows a significant positive correlation for MMLU-Pro, with weaker correlations elsewhere. This highlights a strong model dependency for the effect of openness.

The influence of Conscientiousness also differ across models. In Qwen models, Conscientiousness generally correlated positively with task accuracy, achieving significance on several tasks, partially supporting the human research linking high conscientiousness to better performance. However, in Llama-8B, the correlations are consistently weak, near zero, and non-significant.

Agreeableness shows some positive correlations with accuracy, particularly in Qwen-32B and Llama-8B on several tasks, while being less pronounced in Qwen-14B. Our preliminary analysis of model responses indicates that low Agreeableness scores in Qwen-32B and Llama-8B often correspond to responses showing reluctance to fully address or think through user problems, suggesting reduced cognitive engagement. While Qwen-14B exhibits this reluctance, the length of its chain-of-thought reasoning is not obviously impacted. We hypothesize this variance stems from differences in their training data.

\subsection{Open Tasks}
\begin{figure*}[ht]
    \centering
    \begin{subfigure}[b]{0.32\textwidth}
        \centering
        \includegraphics[width=\textwidth]{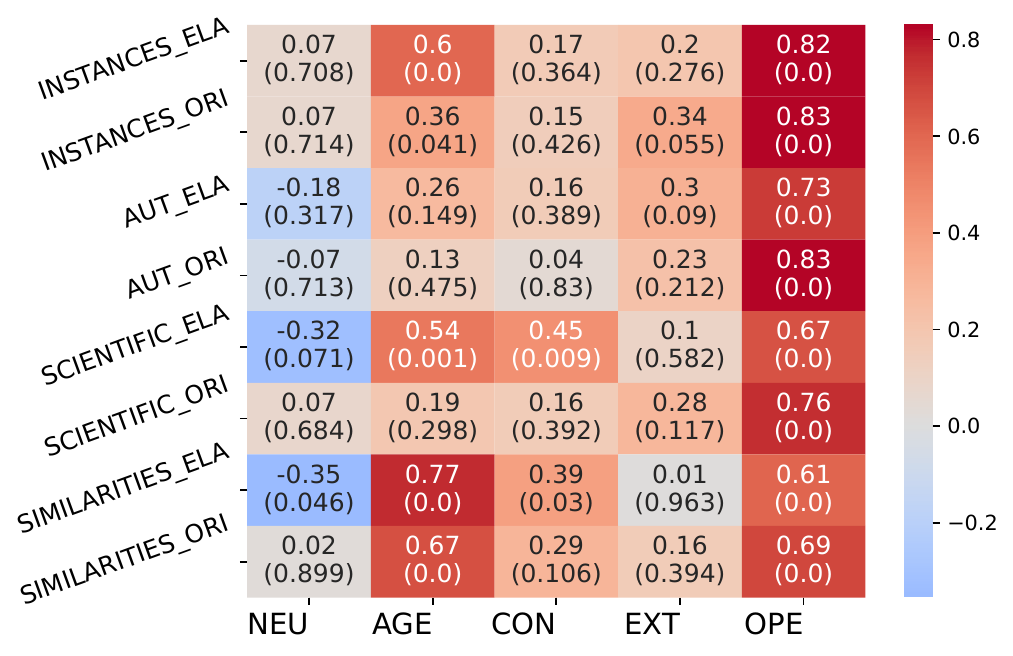}
        \caption{Qwen2.5-32B}
    \end{subfigure}
    \hfill  
    \begin{subfigure}[b]{0.32\textwidth}
        \centering
        \includegraphics[width=\textwidth]{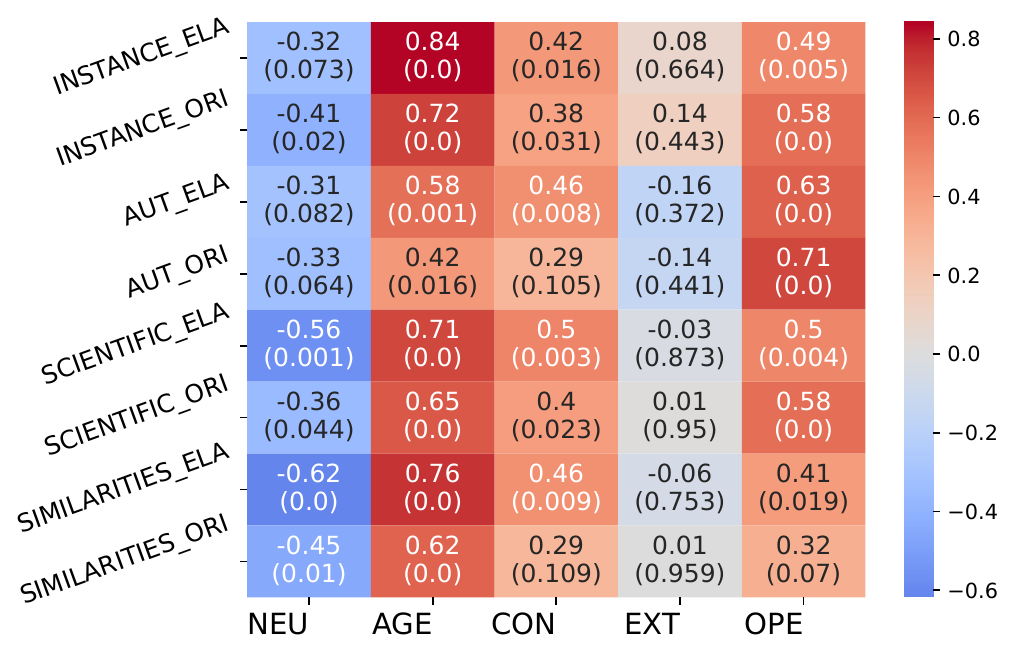}
        \caption{Qwen2.5-14B}
    \end{subfigure}
    \hfill  
    \begin{subfigure}[b]{0.32\textwidth}
        \centering
        \includegraphics[width=\textwidth]{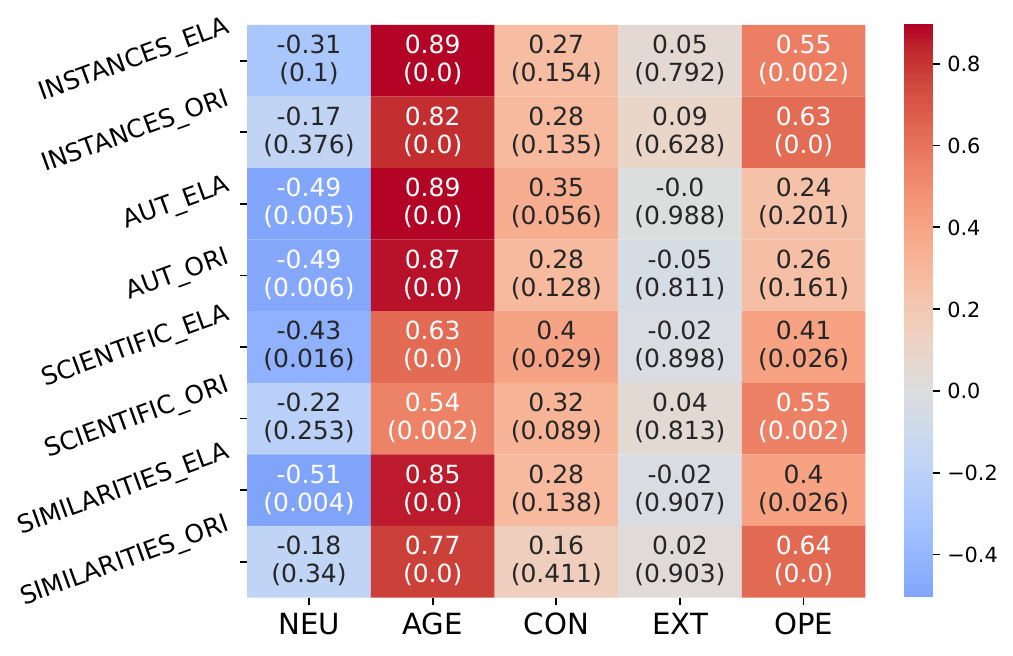}
        \caption{Llama3.1-8B}
    \end{subfigure}
    \caption{Correlation Analysis between Personality Traits and Closed Tasks: Pearson Correlation Coefficients and Statistical Significance (p-values) between Five Personality Dimensions and the Accuracy of Five Closed Tasks.}
    \label{fig:opencor}
\end{figure*}

In Appendix~\ref{s:create}, we present the performance of agents with different Big Five personality trait combinations across the four TTCT evaluation dimensions in the selected creativity tests using different models.
Previous studies have shown that LLMs excel at generating large volumes of output, making the dimensions of fluency and flexibility less important when evaluating their creativity~\cite{DiStefano2024AutomaticSO,Lu2024LLMDE}.
Accordingly, we focus on analyzing the impact of personality traits on originality and elaboration, the two core dimensions of creativity.
\paragraph{Personality and Creativity}
Psychological research indicates that extraversion and openness positively influence human creativity~\cite{KING1996189,Zare2018VoiceCA}. To verify if a similar relationship exists in LLMs and to further investigate how different personality dimensions impact specific aspects of model creativity when simulating human traits,  we calculate the correlation coefficients between each model's personality dimension scores and its demonstrated Originality and Elaboration scores in these tasks, and conduct statistical significance analyses. Detailed results are presented in Figure~\ref{fig:opencor}.

\textbf{Openness }Our results show that Openness has a positive impact on LLM creativity, which is highly consistent with findings from human studies. As shown in Figure~\ref{fig:opencor}, across all models, the Openness score exhibits a significant positive correlation with Originality and Elaboration scores on almost all creativity tasks. This indicates that Openness enhances the models' ability to generate novel and unique ideas and, to some extent, promotes the detailed refinement of these ideas.

\textbf{Extraversion} Unlike findings from human studies, our results don't show a stable, positive impact of Extraversion on LLM's creativity scores. Across all three models and various creativity metrics, the correlation coefficients for Extraversion are mostly small, insignificant, and inconsistent in direction. 

\textbf{Agreeableness} Contrary to findings in some human creativity research suggesting a potential negative correlation for Agreeableness, in this study, the Agreeableness score demonstrate significant positive correlations with both Originality and Elaboration across multiple models and tasks. This result suggests that for LLMs, simulated Agreeableness may contribute to enhancing their scores on these creativity assessment tasks.

\textbf{Neuroticism} The results show a significant negative correlation between the Neuroticism score and multiple creativity metrics. This suggests that lower simulated neuroticism might be more conducive to LLM performance on creativity tasks.

\textbf{Conscientiousness} The relationship between Conscientiousness and creativity metrics shows mixed results. Although significant positive correlations are observed in some model and task combinations, its impact does not exhibit the same cross-model and cross-task consistency as Openness, Agreeableness, or Neuroticism.

\section{Further Influence on Muti-Agents}
This study provides a preliminary exploration of how simulated personality traits influence LLM team collaboration. Given the complexity involved, this serves as an initial attempt to investigate whether these traits affect collaboration.
\subsection{Multi-Agent Team Configuration}
To investigate the impact of simulated personality traits on multi-agent collaboration, we designed different team configurations, detailed in Table~\ref{tab:team_configs}. First, Teams 1-6 have members with identical personality traits to analyze how each Big Five dimension influences collaboration. Second, Teams 7 and 8 include diverse traits to study the effect of personality diversity. Finally, Team 9 introduces a single member with extreme traits to examine their specific impact.
\begin{table}[t] 
    \centering
    \small
    \begin{tabular*}{\columnwidth}{@{\extracolsep{\fill}}llll} 
    \toprule
    \textrm{Team} & \textrm{Member1} & \textrm{Member2} & \textrm{Member3} \\
    \midrule
    Team 1 & \texttt{-{}-{}+++} & \texttt{-{}-{}+++} & \texttt{-{}-{}+++} \\
    Team 2 & \texttt{-{}+-{}++} & \texttt{-{}+-{}++} & \texttt{-{}+-{}++} \\
    Team 3 & \texttt{-{}++-{}+} & \texttt{-{}++-{}+} & \texttt{-{}++-{}+} \\
    Team 4 & \texttt{-{}++++} & \texttt{-{}++++} & \texttt{-{}++++} \\
    Team 5 & \texttt{+++-{}+} & \texttt{+++-{}+} & \texttt{+++-{}+} \\
    Team 6 & \texttt{+++++} & \texttt{+++++} & \texttt{+++++} \\
    \midrule
    Team 7 & \texttt{-{}++++} & \texttt{-{}++-{}+} & \texttt{-{}+-{}++} \\
    Team 8 & \texttt{-{}+++-{}} & \texttt{-{}++++} & \texttt{+++++} \\
    \midrule
    Team 9 & \texttt{-{}++++} & \texttt{+-{}-{}-{}-{}} & \texttt{-{}++++} \\
    \bottomrule
    \end{tabular*} 
    \caption{Personality trait configurations of multi-agent team members.}
    \label{tab:team_configs}
\end{table}

\subsection{Closed Tasks}
Table~\ref{tab:model_benchmarks_teams_close} presents the performance of different teams in closed tasks.
\begin{table}[t]
    \centering
    \small
    \begin{tabular*}{\columnwidth}{@{\extracolsep{\fill}}llccc} 
    \toprule
    \textrm{Model} & \textrm{Team} & \textrm{MMLU} & \textrm{MMLU-Pro} & \textrm{GPQA} \\
    \midrule
    \multirow{9}{*}{Qwen-32B} 
        & Team 1 & 80.87 & 66.71 & 42.06 \\
        & Team 2 & 81.23 & 67.26 & 44.99 \\
        & Team 3 & 81.07 & 67.00 & 44.75 \\
        & Team 4 & 81.48 & 67.23 & 44.93 \\
        & Team 5 & 80.91 & 66.59 & 43.22 \\
        & Team 6 & 81.02 & 67.00 & 42.80 \\
        & Team 7 & 81.43 & 67.11 & 47.25 \\
        
        & Team 8 & 81.37 & 67.30 & 44.44 \\
        & Team 9 & 81.28 & 67.16 & 44.75 \\
    \midrule
    \multirow{9}{*}{Qwen-14B} 
        & Team 1 & 77.89 & 60.94 & 39.38 \\
        & Team 2 & 78.12 & 60.77 & 41.09 \\
        & Team 3 & 77.92 & 60.24 & 38.34 \\
        & Team 4 & 78.02 & 61.20 & 40.29 \\
        & Team 5 & 77.97 & 61.55 & 38.40 \\
        & Team 6 & 78.11 & 61.30 & 37.79 \\
        & Team 7 & 78.13 & 61.10 & 39.56 \\
        
        & Team 8 & 78.25 & 61.73 & 39.44 \\
        & Team 9 & 78.05 & 61.24 & 39.56 \\
    \midrule
    \multirow{9}{*}{llama-8B} 
        & Team 1 & 66.03 & 44.99 & 32.36 \\
        & Team 2 & 67.12 & 45.58 & 31.26 \\
        & Team 3 & 66.60 & 42.82 & 32.78 \\
        & Team 4 & 66.11 & 45.26 & 30.65 \\
        & Team 5 & 65.96 & 44.29 & 30.95 \\
        & Team 6 & 65.84 & 44.03 & 31.38 \\
        & Team 7 & 67.29 & 46.12 & 31.01 \\
        
        & Team 8 & 66.52 & 45.17 & 32.54 \\
        & Team 9 & 66.03 & 45.01 & 30.83 \\
    \bottomrule
    \end{tabular*}
    \caption{Benchmark results on closed tasks across teams for different models.} 
    \label{tab:model_benchmarks_teams_close} 
\end{table}


\begin{table*}[t]
    \centering
    \small  
    \begin{tabular}{@{}llcccccccc@{}}
    \toprule
    \textbf{Model} & \textbf{Team} & \textbf{AUT-O} & \textbf{AUT-E} & \textbf{INS-O} & \textbf{INS-E} & \textbf{SCI-O} & \textbf{SCI-E} & \textbf{SIM-O} & \textbf{SIM-E} \\
    \midrule
    \multirow{10}{*}{Qwen-32B}
        & base & 6.69 & 8.26 & 3.86 & 6.97 & 5.76 & 8.90 & 3.32 & 8.25 \\
        & Team 1 & 7.32 & 8.55 & 3.63 & 6.22 & 5.27 & 8.22 & 2.84 & 4.32 \\
        & Team 2 & 7.19 & 8.40 & 3.87 & 6.96 & 5.57 & 8.40 & 3.72 & 6.43 \\
        & Team 3 & 7.37 & 8.60 & 3.80 & 7.40 & 5.50 & 8.50 & 3.68 & 6.59 \\
        & Team 4 & 7.42 & 8.67 & 3.88 & 7.64 & 5.46 & 8.50 & 3.80 & 7.28 \\
        & Team 5 & 7.41 & 8.60 & 3.66 & 7.31 & 5.36 & 8.31 & 3.08 & 4.14 \\
        & Team 6 & 7.29 & 8.52 & 3.73 & 7.53 & 5.62 & 8.40 & 3.14 & 4.34 \\
        & Team 7 & 7.45 & 8.55 & 3.86 & 7.22 & 5.53 & 8.52 & 3.79 & 6.75 \\
        & Team 8 & 7.38 & 8.55 & 3.80 & 7.40 & 5.39 & 8.43 & 3.46 & 5.94 \\
        & Team 9 & 7.09 & 8.09 & 3.39 & 5.73 & 4.87 & 7.11 & 3.38 & 5.73 \\
        
    \midrule
    \multirow{10}{*}{Qwen-14B}
        & base & 6.08 & 7.46 & 3.55 & 5.77 & 5.46 & 8.70 & 3.51 & 7.86 \\
        & Team 1 & 6.94 & 8.20 & 2.79 & 3.87 & 5.00 & 7.78 & 3.34 & 6.04 \\
        & Team 2 & 6.78 & 8.25 & 3.09 & 4.70 & 5.16 & 8.04 & 3.37 & 6.61 \\
        & Team 3 & 6.90 & 8.30 & 3.10 & 5.03 & 5.22 & 8.29 & 3.44 & 6.93 \\
        & Team 4 & 6.89 & 8.20 & 3.11 & 5.27 & 5.31 & 8.18 & 3.39 & 6.85 \\
        & Team 5 & 6.94 & 8.35 & 2.96 & 5.17 & 5.01 & 7.68 & 3.63 & 6.77 \\
        & Team 6 & 6.61 & 8.06 & 3.20 & 5.23 & 5.11 & 7.83 & 3.54 & 7.18 \\
        & Team 7 & 6.59 & 8.00 & 3.14 & 5.21 & 5.27 & 8.30 & 3.53 & 6.99 \\
        
        & Team 8 & 6.65 & 8.08 & 3.10 & 5.11 & 5.22 & 8.10 & 3.47 & 6.99 \\
        & Team 9 & 6.64 & 7.92 & 2.77 & 4.13 & 4.56 & 6.69 & 3.23 & 6.07 \\
    \midrule
    \multirow{10}{*}{llama-8B}
        & base & 5.04 & 6.87 & 3.30 & 5.71 & 5.04 & 8.24 & 3.32 & 8.25 \\
        & Team 1 & 5.47 & 7.48 & 3.31 & 4.04 & 4.86 & 6.41 & 2.84 & 4.32 \\
        & Team 2 & 5.55 & 7.89 & 3.77 & 6.62 & 5.22 & 7.25 & 3.72 & 6.43 \\
        & Team 3 & 5.62 & 7.60 & 3.78 & 6.09 & 5.14 & 7.57 & 3.68 & 6.59 \\
        & Team 4 & 5.60 & 7.86 & 3.98 & 7.27 & 5.24 & 7.85 & 3.80 & 7.28 \\
        & Team 5 & 4.60 & 5.84 & 3.07 & 3.80 & 4.08 & 5.13 & 3.08 & 4.14 \\
        & Team 6 & 4.77 & 6.01 & 3.28 & 4.41 & 4.52 & 5.84 & 3.14 & 4.34 \\
        & Team 7 & 5.53 & 7.89 & 3.87 & 6.70 & 5.16 & 7.57 & 3.79 & 6.75 \\
        
        & Team 8 & 5.19 & 7.18 & 3.51 & 5.66 & 4.84 & 6.91 & 3.46 & 5.94 \\
        & Team 9 & 5.27 & 6.89 & 3.46 & 5.57 & 4.57 & 6.42 & 3.38 & 5.73 \\
    \bottomrule
    \end{tabular}
    \caption{Benchmark results across teams for different models, covering AUT, INSTANCES (INS), SCIENTIFIC (SCI), and SIMILARITY (SIM). For task, scores are shown for ORIGINALITY (-O) and ELABORATION (-E).}
    \label{tab:model_benchmarks_teams_open} 
\end{table*}


Comparing results in Table~\ref{tab:model_benchmarks_teams_close}, we observe the following trends:
\begin{itemize} 
    \item \textbf{Neuroticism:} High neuroticism (Team 6) generally decreased performance compared to Team 4, indicating a negative impact on collaboration. 
    \item \textbf{Agreeableness:} Low agreeableness (Team 1) consistently performed below Team 4, highlighting the detrimental effect of this trait. 
    \item \textbf{Conscientiousness:} High conscientiousness (Team 2) showed fluctuating results against Team 4, with no clear positive or negative impact in these tasks. 
    \item \textbf{Extraversion:} Low extraversion (Team 3) generally performed worse than Team 4 across most models and tasks. This suggests the potential importance of extraversion for effective collaboration. 
    \item \textbf{Extreme Traits:} Team 9 (with an extreme member) performed slightly below Team 4, but the difference was not significant. In closed tasks, potential negative impacts might be mitigated through internal interactions. 
    \item \textbf{Diversity:} The performance of teams having high diversity (Team 7 and 8) compared to others did not show consistent advantages or disadvantages overall. However, specific advantages were noted on certain models (like Llama-8B) or tasks (e.g., GPQA on Qwen-32B). This suggests diversity may potentially enhance collaboration, particularly for complex problems or with relatively weaker models.
\end{itemize}

\subsection{Open Tasks}
Table \ref{tab:model_benchmarks_teams_open} presents team creativity performance on open tasks. Compared to closed tasks, team composition differences have a greater impact here, with some personality trait configurations showing improved creativity. 

Analyzing individual traits (compared to Team 4): 
\begin{itemize} 
    \item \textbf{Neuroticism:} High neuroticism (Team 6 vs Team 4) generally decreased creativity across most tasks and models. 
    \item \textbf{Agreeableness:} Low agreeableness (Team 1 vs Team 4) resulted in lower performance across most open tasks and models. 
    \item \textbf{Conscientiousness:} Comparing Team 2 (high conscientiousness) and Team 4, results show a slight decrease in creativity scores across most tasks and models, though not always significant.
    \item \textbf{Extraversion:} Low extraversion (Team 3 vs Team 4) generally led to slightly decreased or maintained originality scores across most tasks and models, suggesting higher extraversion might benefit originality in open tasks. 
\end{itemize}

Regarding individuals with extreme traits, Team 9 significantly lagged behind Team 4 across all models and tasks. This gap suggests members with "extreme personality" traits may detrimentally affect team collaboration. Therefore, avoiding extreme personalities during member selection is advisable for building high-performing teams. 

For team diversity, Teams 7 and 8 represent diverse configurations. However, results did not show general or significant advantages for these teams. While achieving good results on some specific tasks, overall performance didn't markedly improve. This suggests mere diversity doesn't consistently enhance team creativity, pointing to the need for precise, task-oriented formation methods in future efforts.

\section{Conclusion}
In this study, we adopts the perspective of "human psychological simulation" to systematically investigate the impact of endowing LLM agents with simulated personality traits, aiming to fill the research gap in this field. We not only examine how simulating different Big Five personality traits affects the performance of individual agents across diverse tasks, but also further analyze the role of these simulated personalities in multi-agent collaborative environments. The results suggest that the abilities of agents are significantly influenced by their simulated personality traits and exhibit characteristics similar to human capabilities. This work provides new evidence and pathways for a deeper understanding and expansion of the capability boundaries of LLMs from the perspective of human simulation.

\section*{Limitations}

\paragraph{The Big Five Personality Traits} \quad The Big Five personality traits exhibit multidimensional complexity, with each dimension continuously distributed between two extremes. This characteristic makes it challenging to comprehensively capture different personality traits and deeply investigate their effects. Therefore, this study focuses on a preliminary exploration of personality traits based on combinations of trait polarities, integrating relevant data such as BFI-2 test scores, and attempts to conduct exploratory research from the perspective of human psychological simulation.

\paragraph{Creativity Assessment}\quad In creativity assessment, we use the Torrance Tests of Creative Thinking (TTCT), which is authoritative in psychology. However, despite being a widely recognized standard for evaluating creativity, TTCT still has some shortcomings. Firstly, the scope of TTCT may not be comprehensive enough to fully encompass creativity. Additionally, its scoring method is relatively subjective, which may lead to some bias in our assessment results.

\section*{Ethics Statement}
Our experiment aims to investigate the impact of simulating human psychological traits, based on the Big Five personality model, on model capabilities. We fully recognize the inherent complexity of personality and acknowledge that it is not possible to perfectly simulate all personality traits. In the evaluation phase, to ensure objectivity and safety, we exclusively use publicly available, common, and non-sensitive test datasets to mitigate the risk of generating harmful or biased results. Our evaluation strictly focuses on the performance changes exhibited by the agent when simulating different personality traits, without involving any content that could potentially cause bias or harm.

\bibliography{anthology,custom}

\appendix
\section{The Big Five Personality Tarits}
The Big Five Personality Traits is a widely recognized model in psychology that summarizes personality traits into five fundamental dimensions~\cite{McCrae1987ValidationOT,Goldberg1990AnA}:

\begin{enumerate}
\item \textbf{Neuroticism}: Also known as emotional instability, this refers to the tendency to experience negative emotions such as anxiety, anger, and depression. Highly neurotic individuals are more possible to anxiety; those low in neuroticism are generally calm, composed, and emotionally stable.
    \item \textbf{Agreeableness}: This dimension reflects the degree of cooperation, empathy, and altruism in interactions with others. Highly agreeable people are friendly, helpful, and trusting; people low in agreeableness may be more suspicious, competitive, and less concerned about others' feelings.

    \item \textbf{Conscientiousness}: This dimension describes the degree of self-discipline, organization, and reliability in goal-directed behavior. Highly conscientious individuals are organized, cautious, responsible, and achievement-oriented; people low in conscientiousness may be more casual, lack planning, and act impulsively.
    \item \textbf{Extraversion}: This refers to the level of energy and need for external stimulation in social interactions. Highly extraverted individuals are enthusiastic, talkative, sociable, and feel energized in groups; those low in extraversion may be quieter, reserved, prefer solitude, and feel energized when alone.
    \item \textbf{Openness}: This refers to an individual's willingness to accept and explore new experiences, ideas, and things. People high in openness are usually imaginative, curious, innovative, and have a strong interest in art; those low in openness tend to be more practical, prefer routines, and have more traditional interests.

\end{enumerate}

Overall, the Big Five Personality Traits model is valued for its scientific validity and universality, making it an important tool for personality assessment in both research and practical applications~\cite{Judge1999THEBF,Gurven2013HowUI}.

Given the prominent status and solid theoretical foundation of the Big Five personality traits model in psychology, this study chooses to base its research on the Big Five model. By simulating the Big Five personality traits, it aims to deeply explore the impact of human psychological simulation on the performance of various model capabilities.
\section{Experiements Configuration}

\label{config}
\subsection{Prompt Conﬁguration}
\label{app: prompts}
We use the following prompt format to shape different Big Five personality traits for the agent. Specifically, we select items that represent the polarity of each dimension to shape the corresponding personality traits.
\begin{tcolorbox}[colframe=gray!70!black, colback=gray!10!white, 
    title=Prompt Template, width=\columnwidth, 
    sharp corners=south, coltitle=black, 
    boxrule=0.8mm]
You are an AI agent designed with specific personality traits based on the Big Five Personality Model. Your behavior, communication style, and decision-making are shaped by the following traits:

1. Neuroticism: ... 

2. Agreeableness: ... 

3. Conscientiousness: ... 

4. Extraversion:  ... 

5. Openness/Intellect: ... 

You must respond according to the described personality traits.
\label{promptset}
\end{tcolorbox}

We use the scale items proposed by ~\citet{DeYoung2007BetweenFA}.Some examples of items are as follows:
\subsection*{Neuroticism (NEU)}
\textbf{Negative :} Rarely get irritated. Keep my emotions under control. Seldom feel blue. \dots \\
\textbf{Positive :} Get angry easily. Get upset easily. Can be stirred up easily. \dots

\subsection*{Agreeableness (AGR)}
\textbf{Negative :} Am not interested in other people's problems. Insult people. \dots \\
\textbf{Positive :} Feel others' emotions. Inquire about others' well-being. Respect authority. \dots

\subsection*{Conscientiousness (CON)}
\textbf{Negative :} Waste my time. Leave my belongings around. \dots \\
\textbf{Positive :} Get things done quickly. Carry out my plans. Like order. Keep things tidy. \dots

\subsection*{Extraversion (EXT)}
\textbf{Negative :} Am hard to get to know. Keep others at a distance. \dots \\
\textbf{Positive :} Warm up quickly to others. Show my feelings when I'm happy. \dots

\subsection*{Openness (OPE)}
\textbf{Negative :} Have difficulty understanding abstract ideas. Seldom get lost in thought. \dots \\
\textbf{Positive :} Am quick to understand things. Like to solve complex problems. \dots

For open tasks, we directly use the above content as the prompt. For closed tasks, we additionally include a format requirement: Answer in the format: 'analysis':'all your thoughts about the question', 'answer':'the number you choose'

\subsection{TTCT}
\label{TTCT}
The Torrance Tests of Creative Thinking (TTCT) is a widely used standardized psychological assessment tool for evaluating an individual's creative potential and thinking characteristics~\cite{torrance1966,Zhao2024AssessingAU}. The scoring of TTCT is mainly based on the following core dimensions:

\begin{itemize}
    \item \textbf{Fluency}: Refers to the number of ideas generated. The more ideas produced, the higher the fluency.
    \item \textbf{Flexibility}: Refers to the diversity of categories of ideas generated. It is the ability to think from different perspectives and come up with different types of ideas.
    \item \textbf{Originality}: Refers to the ability to produce novel, unique, and unusual ideas. The more distinctive and rare the ideas are, the higher the originality is.
    \item \textbf{Elaboration}: Refers to the ability to further refine and add details to the basic ideas, making them more complete.
\end{itemize}

\subsection{Implemention Details}
Our experiment utilizes multiple language models including Qwen2.5-32B-Instruct, Qwen-14B-Instruct, and LLaMA3.1-8B-Instruct, with the temperature set to 0 to ensure deterministic results. We use gpt-4o-mini for creativity assessment through API calls, with the default temperature set to 0.When evaluating the team's performance on closed tasks, we calculate the accuracy by analyzing each member's answers individually. For the evaluation of team creativity, after the discussion, we will assess the creativity of each member’s answers individually. The average of all members’ scores will then be taken as the final team creativity score.

\subsection{Datasets}
GPQA is a dataset specifically designed to evaluate the question-answering capabilities of large language models across challenging, multidisciplinary professional domains~\cite{Rein2023GPQAAG}. It is openly available under a CC BY 4.0 license.

MMLU is a widely used benchmark dataset designed to evaluate the multitask understanding capabilities of large language models~\cite{Hendrycks2020MeasuringMM}, and it is licensed under the MIT License.

MMLU-Pro is a robust and challenging large-scale multitask understanding benchmark that spans multiple domains, designed to provide a stricter evaluation for large language models~\cite{Wang2024MMLUProAM}. It is licensed under the MIT License.

SCIQ is a dataset specifically designed for science question answering, aiming to evaluate the ability models to comprehend scientific facts~\cite{SciQ}. It is licensed under Creative Commons Attribution Non Commercial 3.0.

ARC is a benchmark dataset for scientific question answering, designed to evaluate the understanding and reasoning abilities of artificial intelligence systems~\cite{allenai:arc}. The dataset is openly shared under the Creative Commons Attribution Share Alike 4.0 license.

We follow the work of ~\citet{Lu2024LLMDE} and select the following tests to evaluate creativity.

AUT (Alternative Uses Task) is a classic divergent thinking test~\cite{Wallach1965ModesOT}. In this task, participants are asked to come up with as many novel and unusual uses as possible for a common everyday object.

INSTANCES requires participants to list as many specific examples as possible that belong to a given abstract category or concept~\cite{Wallach1965ModesOT}.

SIMILARITIES typically presents two seemingly unrelated words or concepts and asks participants to identify as many commonalities or similarities between them as possible~\cite{Wallach1965ModesOT}.

The SCIENTIFIC benchmark requires participants to generate as many innovative ideas as possible around scientific topics.

\subsection{LLM Evaluations}
\label{eva}
To ensure the reproducibility and objectivity of evaluations, we reference the evaluation prompts designed by ~\citet{Lu2024LLMDE}. and further optimize some prompts using GPT-4~\cite{openai2023gpt4}. The prompts are shown in Table~\ref{creative:prompt}.

For assessing fluency and flexibility, we instruct the LLM through prompts to eliminate repetitive content and irrelevant responses, followed by quantitative analysis of valid responses. This approach increase scoring bias that might result from simple accumulation of responses. Ultimately, we select the number of responses and category counts as evaluation scores for fluency and flexibility.

For originality and elaboration dimensions, we adopt a holistic evaluation approach to assess the overall innovation level and completeness of responses. We implement a ten-point scoring system with detailed scoring criteria specified in the prompts. Subsequently, we have the LLM conduct comprehensive evaluations of each response to obtain final scores for originality and elaboration.

\begin{table*}
    \centering
    \resizebox{\textwidth}{!}{
    \begin{tabular}{p{53.8pt} p{392.8pt}} 
    \hline 
    \textbf{METRICS} & \textbf{PROMPT} \\ 
    \hline 
    \textbf{Fluency} & 
    You are a thoughtful assistant with a focus on creativity. Identify and count the number of unique, relevant responses and explain why. It is important to the total amount of unique, relevant, and practical responses in the specific format of ``[[X]]'' at the end of your response.\\ 
    \hline 

    \textbf{Flexibility} & 
    You are a helpful assistant and a critical thinker. Please evaluate the flexibility of the relevant responses, where flexibility refers to the variety of distinct categories or perspectives represented in the responses. Define and count the number of unique categories or perspectives present, and provide a brief explanation for how you determined these categories. It is important to present the total number of categories or perspectives in the specific format of ``[[X]]'' at the end of your response.\\ 
    \hline 

    \textbf{Originality} & 
    You are a helpful assistant and a critical thinker. Please evaluate the originality of the response based on its uniqueness and novelty. Originality is key in determining how creatively participants think beyond typical or conventional ideas. Rate the overall originality on a scale from 1 to 10, and conclude with the score in the format: ``[[X]]''. Consider the following guidance:

 1-2 points: Very Common - The idea is mundane and frequently mentioned in everyday contexts. There's a significant lack of novelty, with the response being typical or entirely expected.
 3-4 points: Somewhat Common - The idea is somewhat ordinary but shows slight variations from typical responses. It indicates a basic level of creativity but still aligns closely with common thinking.
 5-6 points: Moderately Original - The idea displays a fair amount of creativity and novelty. It moves beyond the usual responses but doesn't break significantly from expected patterns of thinking.
 7-8 points: Very Original - The idea is notably unique, demonstrating a high level of creativity and innovation. It is unexpected and stands out from more typical ideas.
 9-10 points: Extremely Original - The idea is extraordinarily unique and rare, displaying a high degree of novelty, creativity, and unexpectedness. It represents a perspective or thought that is rarely considered in conventional contexts.

After reviewing these responses, provide an overall originality score based on the above criteria. Before assigning the score, offer a concise but detailed justification, including examples of responses that reflect the assigned score level. Finally, present the overall originality score in the format ``[[X]]''. \\ 
    \hline 
    
    \textbf{Elaboration} & 
    You are a helpful assistant and a critical thinker. Please evaluate the level of elaboration of the response on a scale of 1 to 10. Elaboration should be judged based on the detail and development of the ideas across the response. Conclude with the score in this format: ``[[X]]''. Consider the following guidance:

 1-2 points: Very Basic - The response is extremely basic with minimal detail or explanation. Ideas are presented in a very simple or cursory manner.
 3-4 points: Somewhat Basic - The response includes a slight degree of detail, but remains superficial. Ideas are somewhat developed but lack depth or complexity.
 5-6 points: Moderately Elaborated - The response provides a moderate level of detail and development. Ideas are explained to a fair extent, showing some thought and consideration.
 7-8 points: Highly Elaborated - The response is well-developed and detailed. Ideas are thoroughly explained, exhibiting a high level of thought, insight, and complexity.
 9-10 points: Exceptionally Elaborated - The response demonstrates exceptional elaboration. Ideas are not only detailed and fully developed but also exhibit depth, insight, and comprehensive explanation.

After reviewing these responses, provide an overall elaboration score based on the above criteria. Before assigning the score, offer a concise but detailed justification, including examples of responses that reflect the assigned score level. Finally, present the overall elaboration score in the format ``[[X]]''.\\ 
    \hline 
    \end{tabular}}
    \caption{Evaluation prompts for four different creativity metrics.}
    \label{creative:prompt}
    \end{table*}

\section{Experiments and Results on Single-agent}
\label{Experiment}
\subsection{BFI-2}
\label{BFI-2}
The BFI-2 (Big Five Inventory-2) is a tool used to measure the Big Five personality traits. It provides a more detailed and structurally clearer personality assessment~\cite{Soto2017TheNB}. The BFI-2 assesses the Big Five domains of Extraversion, Agreeableness, Conscientiousness, Neuroticism, and Openness.

Each dimension of the BFI-2 scale includes 3 sub-dimensions (also known as facets), for a total of 15 facets. Each facet consists of 4 items, including 2 positively scored items and 2 negatively scored items, making a total of 60 items in the entire scale.

The scale uses a 5-point Likert scoring method, ranging from "1=Strongly disagree" to "5=Strongly agree". The BFI-2 was developed to enhance the bandwidth, fidelity, and predictive power of personality assessment. It is used in various contexts, including psychological research to explore the relationship between personality traits and behavior, clinical settings to understand personality factors contributing to mental health conditions, and organizational psychology for employee selection, team dynamics, and leadership assessment. Short and extra-short forms of the BFI-2 also exist.

\subsection{BFI-2 on Single-Agent}
We present the BFI-2 test scores of agents with different Big Five personality traits based on various models in Table~\ref{BFI-result-qwen}-\ref{BFI-result-llama}. The test results largely align with our predefined expectations. Additionally, comparisons reveal that models with larger parameter sizes demonstrate a better ability to reflect the designated personality traits in their scores. This finding preliminarily suggests that the capability of a model significantly influences its ability to mimic human traits.

\begin{table}[ht]
\centering
\renewcommand{\arraystretch}{1.4}
\begin{tabular}{lccccc}
\toprule
       Traits & NEU & AGE & CON & EXT & OPE \\
\midrule
    \texttt{-{}-{}-{}-{}-{}} & 1.75 & 2.08 & 2.08 & 1.83 & 1.42 \\
    \texttt{-{}+-{}++} & 1.17 & 5.00 & 3.17 & 3.83 & 4.17 \\
    \texttt{-{}-{}+++} & 1.00 & 2.33 & 4.67 & 4.17 & 3.33 \\
    \texttt{+-{}-{}-{}-{}} & 3.08 & 1.67 & 2.42 & 2.25 & 1.75 \\
    \texttt{-{}-{}++-{}} & 1.75 & 1.83 & 4.67 & 4.25 & 1.17 \\
    \texttt{-{}-{}-{}+-{}} & 1.42 & 2.00 & 2.83 & 3.75 & 1.33 \\
    \texttt{++-{}++} & 4.50 & 3.08 & 2.25 & 3.83 & 4.17 \\
    \texttt{-{}+-{}-{}+} & 1.58 & 4.08 & 2.67 & 2.25 & 4.33 \\
    \texttt{-{}++++} & 1.42 & 4.92 & 4.92 & 4.17 & 4.67 \\
    \texttt{-{}++-{}+} & 1.42 & 4.75 & 4.67 & 2.42 & 4.08 \\
    \texttt{+-{}-{}++} & 4.75 & 1.33 & 1.83 & 4.58 & 3.92 \\
    \texttt{-{}+-{}-{}-{}} & 1.75 & 3.75 & 3.17 & 1.75 & 2.25 \\
    \texttt{++-{}+-{}} & 4.67 & 3.17 & 2.17 & 3.83 & 2.17 \\
    \texttt{+-{}-{}+-{}} & 4.83 & 1.25 & 1.58 & 4.17 & 1.75 \\
    \texttt{+-{}+-{}-{}} & 3.50 & 1.50 & 3.92 & 1.50 & 1.58 \\
    \texttt{+-{}+++} & 4.75 & 1.33 & 4.25 & 4.25 & 4.17 \\
    \texttt{-{}-{}-{}++} & 1.17 & 2.17 & 2.33 & 4.00 & 2.08 \\
    \texttt{++-{}-{}+} & 4.50 & 3.00 & 2.00 & 1.75 & 3.92 \\
    \texttt{+-{}-{}-{}+} & 4.33 & 1.25 & 1.83 & 2.08 & 3.75 \\
    \texttt{+++++} & 4.75 & 3.08 & 4.58 & 3.83 & 4.17 \\
    \texttt{++-{}-{}-{}} & 4.42 & 2.67 & 2.25 & 2.17 & 2.50 \\
    \texttt{-{}-{}+-{}+} & 1.17 & 2.08 & 4.50 & 2.33 & 3.08 \\
    \texttt{+++-{}+} & 4.33 & 3.33 & 4.17 & 2.17 & 3.92 \\
    \texttt{-{}-{}-{}-{}+} & 1.50 & 1.67 & 2.17 & 2.08 & 3.33 \\
    \texttt{-{}++-{}-{}} & 1.67 & 4.00 & 4.42 & 2.08 & 2.25 \\
    \texttt{+++-{}-{}} & 3.58 & 3.00 & 3.92 & 2.08 & 2.08 \\
    \texttt{++++-{}} & 4.58 & 2.83 & 4.33 & 3.42 & 2.25 \\
    \texttt{+-{}+-{}+} & 3.83 & 1.50 & 4.08 & 2.25 & 3.67 \\
    \texttt{+-{}++-{}} & 4.25 & 1.75 & 4.08 & 4.33 & 2.08 \\
    \texttt{-{}+-{}+-{}} & 1.58 & 4.67 & 3.25 & 3.75 & 2.42 \\
    \texttt{-{}+++-{}} & 1.67 & 4.75 & 4.92 & 4.08 & 2.25 \\
    \texttt{-{}-{}+-{}-{}} & 1.75 & 1.92 & 4.75 & 2.08 & 1.58 \\
\bottomrule
\end{tabular}
\caption{BFI-2 Scores of Agents Configured with Different Big Five Personality Profiles On Llama1.3-8B.}
\label{BFI-result-llama}
\end{table}

\begin{table*}[ht]
\centering
\renewcommand{\arraystretch}{1.4}
\begin{subtable}[t]{0.45\textwidth}
\centering
\begin{tabular}{lccccc}
\toprule
       Traits & NEU & AGE & CON & EXT & OPE \\
\midrule
    \texttt{-{}-{}-{}-{}-{}} & 1.33 & 1.17 & 1.25 & 1.67 & 1.08 \\
    \texttt{-{}+-{}++} & 1.00 & 4.92 & 2.17 & 4.00 & 4.67 \\
    \texttt{-{}-{}+++} & 1.00 & 1.17 & 5.00 & 5.00 & 3.67 \\
    \texttt{+-{}-{}-{}-{}} & 5.00 & 1.00 & 1.17 & 1.58 & 1.08 \\
    \texttt{-{}-{}++-{}} & 1.00 & 1.17 & 5.00 & 4.92 & 1.17 \\
    \texttt{-{}-{}-{}+-{}} & 1.08 & 1.08 & 1.25 & 4.75 & 1.25 \\
    \texttt{++-{}++} & 4.92 & 4.33 & 1.25 & 3.75 & 4.75 \\
    \texttt{-{}+-{}-{}+} & 1.00 & 4.83 & 2.50 & 1.58 & 4.92 \\
    \texttt{-{}++++} & 1.00 & 4.92 & 5.00 & 4.33 & 5.00 \\
    \texttt{-{}++-{}+} & 1.00 & 4.92 & 5.00 & 2.25 & 4.83 \\
    \texttt{+-{}-{}++} & 4.92 & 1.00 & 1.17 & 4.83 & 4.92 \\
    \texttt{-{}+-{}-{}-{}} & 1.00 & 4.67 & 2.42 & 1.50 & 1.58 \\
    \texttt{++-{}+-{}} & 4.92 & 4.08 & 1.25 & 3.75 & 1.50 \\
    \texttt{+-{}-{}+-{}} & 5.00 & 1.00 & 1.17 & 4.75 & 1.17 \\
    \texttt{+-{}+-{}-{}} & 5.00 & 1.00 & 5.00 & 1.83 & 1.00 \\
    \texttt{+-{}+++} & 5.00 & 1.00 & 5.00 & 5.00 & 4.92 \\
    \texttt{-{}-{}-{}++} & 1.00 & 1.08 & 1.25 & 5.00 & 2.25 \\
    \texttt{++-{}-{}+} & 5.00 & 3.58 & 1.25 & 1.42 & 4.67 \\
    \texttt{+-{}-{}-{}+} & 5.00 & 1.00 & 1.17 & 1.58 & 3.75 \\
    \texttt{+++++} & 4.67 & 4.08 & 4.92 & 4.17 & 4.92 \\
    \texttt{++-{}-{}-{}} & 5.00 & 3.58 & 1.25 & 1.50 & 1.33 \\
    \texttt{-{}-{}+-{}+} & 1.00 & 1.42 & 5.00 & 2.17 & 4.08 \\
    \texttt{+++-{}+} & 4.92 & 3.83 & 4.92 & 1.75 & 4.75 \\
    \texttt{-{}-{}-{}-{}+} & 1.08 & 1.08 & 1.25 & 1.75 & 2.83 \\
    \texttt{-{}++-{}-{}} & 1.00 & 4.92 & 5.00 & 2.17 & 2.08 \\
    \texttt{+++-{}-{}} & 5.00 & 4.00 & 4.67 & 1.67 & 1.25 \\
    \texttt{++++-{}} & 4.92 & 4.25 & 4.92 & 4.17 & 1.92 \\
    \texttt{+-{}+-{}+} & 5.00 & 1.00 & 5.00 & 2.00 & 4.75 \\
    \texttt{+-{}++-{}} & 5.00 & 1.00 & 5.00 & 4.92 & 1.08 \\
    \texttt{-{}+-{}+-{}} & 1.00 & 4.92 & 2.33 & 3.92 & 2.42 \\
    \texttt{-{}+++-{}} & 1.00 & 4.92 & 5.00 & 4.25 & 2.50 \\
    \texttt{-{}-{}+-{}-{}} & 1.00 & 1.17 & 5.00 & 2.08 & 1.17 \\
\bottomrule
\end{tabular}
\caption{Qwen2.5-32B-Instruct}
\end{subtable}
\hfill
\begin{subtable}[t]{0.45\textwidth}
\centering
\begin{tabular}{lccccc}
\toprule
       Traits & NEU & AGE & CON & EXT & OPE \\
\midrule
    \texttt{-{}-{}-{}-{}-{}} & 1.00 & 1.50 & 2.17 & 2.17 & 1.75 \\
    \texttt{-{}+-{}++} & 1.00 & 4.83 & 2.67 & 4.00 & 4.33 \\
    \texttt{-{}-{}+++} & 1.08 & 2.75 & 4.92 & 4.25 & 4.33 \\
    \texttt{+-{}-{}-{}-{}} & 4.92 & 1.17 & 1.17 & 2.25 & 1.75 \\
    \texttt{-{}-{}++-{}} & 1.17 & 2.58 & 5.00 & 3.92 & 1.67 \\
    \texttt{-{}-{}-{}+-{}} & 1.08 & 1.92 & 1.75 & 3.83 & 1.92 \\
    \texttt{++-{}++} & 4.83 & 4.08 & 2.00 & 4.00 & 4.33 \\
    \texttt{-{}+-{}-{}+} & 1.08 & 4.75 & 3.00 & 1.92 & 4.58 \\
    \texttt{-{}++++} & 1.00 & 4.75 & 5.00 & 3.67 & 5.00 \\
    \texttt{-{}++-{}+} & 1.00 & 4.92 & 4.92 & 2.42 & 4.33 \\
    \texttt{+-{}-{}++} & 4.75 & 1.08 & 1.42 & 4.00 & 3.67 \\
    \texttt{-{}+-{}-{}-{}} & 1.17 & 4.83 & 2.67 & 1.92 & 2.25 \\
    \texttt{++-{}+-{}} & 4.67 & 3.83 & 1.83 & 3.83 & 2.33 \\
    \texttt{+-{}-{}+-{}} & 4.83 & 1.25 & 1.42 & 4.00 & 1.83 \\
    \texttt{+-{}+-{}-{}} & 4.83 & 1.00 & 3.50 & 2.33 & 2.08 \\
    \texttt{+-{}+++} & 4.75 & 1.42 & 3.67 & 4.58 & 4.00 \\
    \texttt{-{}-{}-{}++} & 1.08 & 1.75 & 2.00 & 4.25 & 3.08 \\
    \texttt{++-{}-{}+} & 4.75 & 3.83 & 1.50 & 1.92 & 4.17 \\
    \texttt{+-{}-{}-{}+} & 5.00 & 1.00 & 1.33 & 1.83 & 3.67 \\
    \texttt{+++++} & 4.67 & 4.17 & 4.67 & 3.83 & 4.67 \\
    \texttt{++-{}-{}-{}} & 4.92 & 3.67 & 1.50 & 1.67 & 2.08 \\
    \texttt{-{}-{}+-{}+} & 1.08 & 2.33 & 5.00 & 2.08 & 4.08 \\
    \texttt{+++-{}+} & 4.75 & 3.92 & 4.58 & 2.08 & 4.50 \\
    \texttt{-{}-{}-{}-{}+} & 1.08 & 1.75 & 2.42 & 2.00 & 3.83 \\
    \texttt{-{}++-{}-{}} & 1.08 & 4.75 & 4.83 & 2.08 & 2.00 \\
    \texttt{+++-{}-{}} & 4.83 & 3.58 & 4.75 & 1.92 & 2.08 \\
    \texttt{++++-{}} & 4.67 & 3.75 & 4.58 & 4.00 & 2.17 \\
    \texttt{+-{}+-{}+} & 4.92 & 1.00 & 3.92 & 2.42 & 4.42 \\
    \texttt{+-{}++-{}} & 4.75 & 1.17 & 3.83 & 4.25 & 1.67 \\
    \texttt{-{}+-{}+-{}} & 1.08 & 4.75 & 3.00 & 2.83 & 2.42 \\
    \texttt{-{}+++-{}} & 1.08 & 4.75 & 5.00 & 3.33 & 2.33 \\
    \texttt{-{}-{}+-{}-{}} & 1.08 & 2.25 & 5.00 & 2.33 & 1.67 \\
\bottomrule
\end{tabular}
\caption{Qwen2.5-14B-Instruct}
\end{subtable}
\caption{BFI-2 Scores of Language Model Agents Configured with Different Big Five Personality Profiles.}
\label{BFI-result-qwen}
\end{table*}

\subsection{Single-Agent Performance}
\label{apd:resultofclose}
In Table~\ref{tab:qwen32-close}-\ref{tab:llama-close}, we present detailed performance results of agents with different Big Five personality traits in closed tasks. The data demonstrates that there are significant correlations between agents' personality trait differences and their task completion capabilities. Agents with different personality traits exhibit distinct performance levels when handling closed tasks.

\begin{table*}
    \centering
    
    \resizebox{\textwidth}{!}{
    \begin{tabular}{lcccccc}
    \toprule
           Trait &  MMLU(\%) &  MMLU-Pro(\%) &  SCIQ(\%) &  ARC-Challenge(\%) &  ARC-Easy(\%) &  AVERAGE(\%) \\
    \midrule
    \texttt{-{}++++} & 80.37 & 61.6 & 95.7 & 94.97 & 98.53 & 86.23 \\
    \texttt{-{}++-{}+} & 80.2 & 61.36 & 95.7 & 94.45 & 98.57 & 86.06 \\
    \texttt{-{}+-{}++} & 80.39 & 61.09 & 95.3 & 94.88 & 98.4 & 86.01 \\
    \texttt{-{}-{}+-{}+} & 79.64 & 61.1 & 96.0 & 94.28 & 98.48 & 85.9 \\
    \texttt{-{}-{}+++} & 79.76 & 60.9 & 95.2 & 94.8 & 98.15 & 85.76 \\
    \texttt{-{}+-{}-{}+} & 79.74 & 60.75 & 95.2 & 94.54 & 98.48 & 85.74 \\
    \texttt{-{}-{}-{}-{}+} & 79.65 & 59.52 & 95.7 & 94.71 & 98.27 & 85.57 \\
    \texttt{+++++} & 79.37 & 59.08 & 95.7 & 94.62 & 98.32 & 85.42 \\
    \texttt{-{}++-{}-{}} & 79.51 & 59.92 & 95.1 & 94.03 & 98.32 & 85.38 \\
    \texttt{-{}-{}++-{}} & 78.92 & 59.57 & 95.7 & 94.62 & 98.06 & 85.37 \\
    \texttt{-{}-{}-{}++} & 79.32 & 59.87 & 94.7 & 94.8 & 98.15 & 85.37 \\
    \texttt{-{}+++-{}} & 79.35 & 59.41 & 95.7 & 94.28 & 98.06 & 85.36 \\
    \texttt{-{}+-{}+-{}} & 79.28 & 58.75 & 95.4 & 94.71 & 98.32 & 85.29 \\
    \texttt{+++-{}+} & 78.97 & 58.65 & 95.6 & 94.37 & 98.57 & 85.23 \\
    \texttt{+-{}+++} & 79.01 & 58.98 & 95.3 & 94.54 & 98.32 & 85.23 \\
    \texttt{+-{}+-{}+} & 78.68 & 58.16 & 95.3 & 93.77 & 98.57 & 84.9 \\
    \texttt{-{}+-{}-{}-{}} & 78.59 & 58.14 & 94.8 & 94.45 & 98.06 & 84.81 \\
    \texttt{++-{}++} & 78.54 & 57.65 & 95.2 & 94.28 & 98.4 & 84.81 \\
    \texttt{-{}-{}+-{}-{}} & 78.8 & 58.05 & 94.7 & 93.94 & 98.11 & 84.72 \\
    \texttt{++++-{}} & 78.34 & 57.1 & 95.0 & 94.71 & 98.19 & 84.67 \\
    \texttt{++-{}-{}+} & 78.61 & 56.91 & 95.1 & 94.2 & 98.36 & 84.64 \\
    \texttt{+-{}-{}-{}+} & 78.3 & 56.39 & 95.8 & 94.37 & 98.02 & 84.58 \\
    \texttt{+-{}-{}++} & 78.55 & 57.45 & 94.3 & 94.2 & 97.81 & 84.46 \\
    \texttt{+++-{}-{}} & 77.66 & 55.37 & 95.2 & 93.94 & 98.4 & 84.11 \\
    \texttt{+-{}++-{}} & 77.35 & 55.39 & 94.4 & 93.77 & 97.98 & 83.78 \\
    \texttt{++-{}+-{}} & 77.46 & 55.26 & 94.4 & 93.86 & 97.73 & 83.74 \\
    \texttt{+-{}+-{}-{}} & 76.61 & 54.78 & 94.8 & 93.6 & 97.39 & 83.44 \\
    \texttt{++-{}-{}-{}} & 76.76 & 54.26 & 94.2 & 94.11 & 97.81 & 83.43 \\
    \texttt{-{}-{}-{}+-{}} & 76.24 & 54.42 & 94.8 & 93.43 & 97.26 & 83.23 \\
    \texttt{-{}-{}-{}-{}-{}} & 76.1 & 53.13 & 94.8 & 92.75 & 97.73 & 82.9 \\
    \texttt{+-{}-{}+-{}} & 75.53 & 52.21 & 94.3 & 93.17 & 97.14 & 82.47 \\
    \texttt{+-{}-{}-{}-{}} & 73.83 & 50.32 & 94.1 & 91.81 & 96.84 & 81.38 \\
    \bottomrule
\end{tabular}}
\caption{Accuracy of Agents with Different Big Five Personality Traits on Closed Tasks using Qwen2.5-32B-Instruct.}
\label{tab:qwen32-close}
\end{table*}

\begin{table*}
    \centering
    
    \resizebox{\textwidth}{!}{
    \begin{tabular}{lcccccc}
    \toprule
           Trait &  MMLU(\%) &  MMLU-Pro(\%) &  SCIQ(\%) &  ARC-Challenge(\%) &  ARC-Easy(\%) &  AVERAGE(\%) \\
    \midrule
    \texttt{-{}++++} & 73.45 & 51.5 & 94.2 & 89.59 & 96.04 & 80.96 \\
    \texttt{-{}+++-{}} & 73.34 & 50.66 & 94.3 & 90.19 & 96.21 & 80.94 \\
    \texttt{-{}++-{}-{}} & 73.69 & 50.82 & 94.2 & 90.1 & 95.79 & 80.92 \\
    \texttt{-{}-{}-{}-{}+} & 73.59 & 50.44 & 94.0 & 89.85 & 96.68 & 80.91 \\
    \texttt{-{}+-{}-{}-{}} & 73.69 & 50.96 & 94.0 & 89.85 & 95.96 & 80.89 \\
    \texttt{-{}+-{}+-{}} & 73.76 & 51.71 & 93.7 & 89.33 & 95.96 & 80.89 \\
    \texttt{-{}-{}+-{}-{}} & 73.71 & 50.12 & 94.4 & 89.85 & 96.38 & 80.89 \\
    \texttt{-{}+-{}-{}+} & 73.33 & 51.17 & 94.2 & 89.85 & 95.88 & 80.89 \\
    \texttt{-{}-{}++-{}} & 73.29 & 50.07 & 94.3 & 90.44 & 96.3 & 80.88 \\
    \texttt{-{}-{}-{}++} & 73.89 & 50.28 & 94.3 & 89.68 & 96.21 & 80.87 \\
    \texttt{-{}-{}+++} & 73.41 & 51.02 & 94.2 & 89.42 & 96.25 & 80.86 \\
    \texttt{-{}+-{}++} & 73.52 & 51.15 & 93.5 & 90.1 & 95.83 & 80.82 \\
    \texttt{-{}-{}+-{}+} & 73.36 & 50.22 & 93.8 & 90.1 & 96.59 & 80.81 \\
    \texttt{-{}++-{}+} & 73.41 & 51.14 & 93.9 & 88.82 & 96.3 & 80.71 \\
    \texttt{-{}-{}-{}-{}-{}} & 73.31 & 49.89 & 93.6 & 89.93 & 96.13 & 80.57 \\
    \texttt{-{}-{}-{}+-{}} & 73.27 & 49.24 & 94.1 & 89.51 & 96.13 & 80.45 \\
    \texttt{+++-{}+} & 73.27 & 49.82 & 93.7 & 89.42 & 95.83 & 80.41 \\
    \texttt{+++++} & 72.53 & 50.29 & 93.7 & 88.99 & 95.5 & 80.2 \\
    \texttt{++++-{}} & 72.45 & 49.16 & 94.5 & 89.42 & 95.24 & 80.15 \\
    \texttt{++-{}++} & 72.75 & 50.07 & 93.4 & 88.57 & 95.66 & 80.09 \\
    \texttt{+-{}+++} & 72.77 & 48.3 & 94.4 & 88.74 & 96.17 & 80.08 \\
    \texttt{+++-{}-{}} & 72.69 & 48.3 & 94.1 & 89.08 & 95.75 & 79.98 \\
    \texttt{++-{}-{}+} & 72.51 & 48.75 & 93.8 & 88.65 & 95.75 & 79.89 \\
    \texttt{+-{}+-{}+} & 72.21 & 48.25 & 94.0 & 88.91 & 95.92 & 79.86 \\
    \texttt{++-{}+-{}} & 72.13 & 48.03 & 94.3 & 88.82 & 96.0 & 79.86 \\
    \texttt{+-{}-{}-{}+} & 71.98 & 47.81 & 92.6 & 89.68 & 95.83 & 79.58 \\
    \texttt{+-{}++-{}} & 71.96 & 47.28 & 93.6 & 89.42 & 95.58 & 79.57 \\
    \texttt{+-{}-{}++} & 72.23 & 48.49 & 93.2 & 88.23 & 95.54 & 79.54 \\
    \texttt{+-{}+-{}-{}} & 71.57 & 46.1 & 93.7 & 89.33 & 95.62 & 79.26 \\
    \texttt{++-{}-{}-{}} & 71.48 & 47.12 & 93.2 & 88.23 & 95.88 & 79.18 \\
    \texttt{+-{}-{}+-{}} & 70.94 & 45.35 & 92.9 & 88.99 & 95.83 & 78.8 \\
    \texttt{+-{}-{}-{}-{}} & 70.71 & 44.21 & 92.7 & 88.31 & 95.03 & 78.19 \\
    \bottomrule
\end{tabular}}
\caption{Accuracy of Agents with Different Big Five Personality Traits on Closed Tasks using Qwen2.5-14B-Instruct.}
\label{tab:qwen14-close}
\end{table*}

\begin{table*}
    \centering

    \resizebox{\textwidth}{!}{
    \begin{tabular}{lcccccc}
    \toprule
           Trait &  MMLU(\%) &  MMLU-Pro(\%) &  SCIQ(\%) &  ARC-Challenge(\%) &  ARC-Easy(\%) &  AVERAGE(\%) \\
    \midrule
    \texttt{++-{}-{}+} & 62.05 & 37.36 & 89.10 & 79.44 & 89.31 & 71.45 \\
    \texttt{++-{}++} & 61.84 & 37.46 & 88.40 & 79.35 & 89.60 & 71.33 \\
    \texttt{-{}-{}-{}+-{}} & 61.86 & 35.26 & 88.40 & 78.84 & 90.45 & 70.96 \\
    \texttt{-{}-{}-{}-{}+} & 61.67 & 35.68 & 88.60 & 78.24 & 90.45 & 70.93 \\
    \texttt{+++-{}-{}} & 61.84 & 36.32 & 87.30 & 79.52 & 89.56 & 70.91 \\
    \texttt{-{}-{}+++} & 61.64 & 35.15 & 88.30 & 78.92 & 90.45 & 70.89 \\
    \texttt{-{}+-{}-{}-{}} & 61.67 & 34.75 & 88.20 & 79.01 & 90.70 & 70.87 \\
    \texttt{++-{}+-{}} & 61.22 & 37.13 & 88.10 & 78.75 & 89.06 & 70.85 \\
    \texttt{-{}++-{}-{}} & 61.55 & 35.33 & 88.80 & 77.73 & 90.61 & 70.80 \\
    \texttt{-{}+-{}-{}+} & 61.76 & 35.64 & 88.10 & 77.47 & 90.70 & 70.73 \\
    \texttt{+++-{}+} & 61.81 & 36.64 & 88.30 & 77.39 & 89.18 & 70.66 \\
    \texttt{-{}-{}+-{}+} & 61.21 & 35.18 & 88.80 & 77.99 & 89.90 & 70.62 \\
    \texttt{++++-{}} & 60.98 & 36.88 & 87.90 & 78.41 & 88.68 & 70.57 \\
    \texttt{-{}+-{}+-{}} & 61.12 & 35.00 & 88.00 & 78.16 & 90.11 & 70.48 \\
    \texttt{-{}++-{}+} & 61.49 & 35.62 & 87.10 & 79.01 & 89.06 & 70.46 \\
    \texttt{+++++} & 60.90 & 37.31 & 88.40 & 77.56 & 87.88 & 70.41 \\
    \texttt{-{}-{}-{}++} & 61.53 & 34.95 & 88.00 & 77.56 & 89.98 & 70.40 \\
    \texttt{-{}-{}++-{}} & 61.23 & 34.36 & 88.10 & 77.82 & 90.49 & 70.40 \\
    \texttt{++-{}-{}-{}} & 61.69 & 36.76 & 87.70 & 76.62 & 88.89 & 70.33 \\
    \texttt{-{}+++-{}} & 61.17 & 35.04 & 87.40 & 77.65 & 89.81 & 70.21 \\
    \texttt{-{}++++} & 60.74 & 35.10 & 87.30 & 77.82 & 89.60 & 70.11 \\
    \texttt{-{}+-{}++} & 61.15 & 35.46 & 87.20 & 77.13 & 89.27 & 70.04 \\
    \texttt{-{}-{}-{}-{}-{}} & 60.63 & 34.42 & 87.50 & 77.13 & 89.60 & 69.86 \\
    \texttt{-{}-{}+-{}-{}} & 59.62 & 33.57 & 87.40 & 76.88 & 89.48 & 69.39 \\
    \texttt{+-{}-{}-{}+} & 57.86 & 35.57 & 87.70 & 74.66 & 87.54 & 68.67 \\
    \texttt{+-{}+++} & 58.00 & 35.65 & 87.70 & 73.55 & 87.54 & 68.49 \\
    \texttt{+-{}-{}++} & 58.49 & 35.70 & 86.40 & 74.06 & 86.83 & 68.30 \\
    \texttt{+-{}++-{}} & 57.11 & 34.65 & 87.20 & 75.00 & 87.33 & 68.26 \\
    \texttt{+-{}-{}+-{}} & 57.33 & 34.08 & 87.60 & 74.49 & 87.46 & 68.19 \\
    \texttt{+-{}-{}-{}-{}} & 57.61 & 34.08 & 87.30 & 73.72 & 88.09 & 68.16 \\
    \texttt{+-{}+-{}+} & 57.62 & 35.14 & 88.30 & 71.84 & 86.07 & 67.79 \\
    \texttt{+-{}+-{}-{}} & 57.09 & 34.07 & 86.60 & 74.23 & 86.49 & 67.70 \\
    \bottomrule
    \end{tabular}}
    \caption{Accuracy of Agents with Different Big Five Personality Traits on Closed Tasks using llama3.1-8B.}
    \label{tab:llama-close}
\end{table*}

\begin{table*}
\centering

\begin{tabular}{lcccccccc}
\hline
\multirow{2}{*}{Trait} & \multicolumn{2}{c}{ORIGINALITY} & \multicolumn{2}{c}{ELABORATION} & \multicolumn{2}{c}{FLUENCY} & \multicolumn{2}{c}{FLEXIBILITY} \\
\cline{2-9}
 & Mean & Std. & Mean & Std. & Mean & Std. & Mean & Std. \\
\hline
\texttt{-{}++-{}+} & 7.48 & 0.71 & 8.41 & 0.71 & 12.7 & 1.92 & 7.48 & 1.68 \\
\texttt{+-{}-{}++} & 7.51 & 0.74 & 8.38 & 0.71 & 11.99 & 2.2 & 7.15 & 1.44 \\
\texttt{+-{}+-{}+} & 7.05 & 0.97 & 8.22 & 0.61 & 11.17 & 1.84 & 7.0 & 1.56 \\
\texttt{-{}++++} & 7.01 & 0.85 & 8.16 & 0.69 & 13.48 & 1.96 & 7.61 & 1.71 \\
\texttt{++-{}-{}+} & 7.06 & 0.83 & 8.05 & 0.75 & 11.32 & 1.9 & 6.87 & 1.65 \\
\texttt{-{}+-{}-{}+} & 6.99 & 0.93 & 8.12 & 0.64 & 11.81 & 2.12 & 7.13 & 1.76 \\
\texttt{+++-{}+} & 6.89 & 0.99 & 8.18 & 0.65 & 11.56 & 1.92 & 6.86 & 1.76 \\
\texttt{+++++} & 6.95 & 0.85 & 8.12 & 0.64 & 11.75 & 2.12 & 7.08 & 1.62 \\
\texttt{-{}-{}-{}-{}+} & 7.0 & 1.01 & 8.03 & 0.71 & 11.85 & 2.04 & 7.11 & 1.84 \\
\texttt{+-{}+++} & 6.89 & 0.95 & 8.02 & 0.63 & 11.36 & 1.87 & 7.26 & 1.84 \\
\texttt{-{}-{}+-{}+} & 6.8 & 0.89 & 8.03 & 0.67 & 11.41 & 2.11 & 6.99 & 2.02 \\
\texttt{-{}-{}+++} & 6.77 & 0.8 & 8.04 & 0.62 & 12.74 & 2.04 & 7.33 & 1.91 \\
\texttt{++-{}++} & 6.8 & 0.87 & 7.99 & 0.69 & 11.5 & 2.16 & 7.11 & 1.65 \\
\texttt{-{}+-{}++} & 6.68 & 0.84 & 8.02 & 0.65 & 12.81 & 2.09 & 7.44 & 1.78 \\
\texttt{+-{}-{}-{}+} & 6.79 & 0.96 & 7.9 & 1.03 & 11.56 & 2.14 & 7.2 & 1.71 \\
\texttt{-{}+++-{}} & 6.53 & 1.06 & 7.81 & 0.77 & 12.72 & 2.21 & 7.4 & 1.79 \\
\texttt{-{}-{}-{}++} & 6.52 & 1.07 & 7.79 & 0.71 & 12.31 & 2.28 & 7.15 & 1.82 \\
\texttt{-{}+-{}+-{}} & 6.4 & 0.92 & 7.85 & 0.79 & 12.34 & 2.17 & 7.23 & 1.96 \\
\texttt{-{}-{}-{}+-{}} & 6.59 & 0.94 & 7.66 & 0.78 & 11.97 & 2.19 & 7.43 & 2.09 \\
\texttt{++++-{}} & 6.43 & 0.99 & 7.79 & 0.7 & 11.75 & 2.15 & 7.31 & 1.86 \\
\texttt{-{}-{}++-{}} & 6.42 & 0.9 & 7.73 & 0.76 & 12.08 & 2.19 & 7.08 & 1.79 \\
\texttt{-{}++-{}-{}} & 6.39 & 0.94 & 7.71 & 0.88 & 11.93 & 1.99 & 7.43 & 1.9 \\
\texttt{+-{}-{}+-{}} & 6.42 & 0.99 & 7.64 & 0.84 & 11.48 & 2.14 & 7.37 & 1.91 \\
\texttt{++-{}+-{}} & 6.43 & 0.96 & 7.63 & 0.88 & 11.11 & 1.84 & 7.11 & 1.75 \\
\texttt{+-{}++-{}} & 6.41 & 0.88 & 7.64 & 0.83 & 11.55 & 2.11 & 7.4 & 1.96 \\
\texttt{-{}-{}-{}-{}-{}} & 6.23 & 1.24 & 7.35 & 1.03 & 11.15 & 1.88 & 6.85 & 1.63 \\
\texttt{-{}+-{}-{}-{}} & 6.1 & 1.03 & 7.4 & 0.97 & 11.13 & 1.88 & 6.98 & 1.75 \\
\texttt{++-{}-{}-{}} & 6.17 & 0.94 & 7.31 & 0.97 & 10.65 & 1.42 & 6.94 & 1.66 \\
\texttt{+++-{}-{}} & 5.84 & 0.93 & 7.31 & 0.87 & 10.87 & 1.53 & 7.02 & 1.86 \\
\texttt{-{}-{}+-{}-{}} & 5.83 & 0.95 & 7.2 & 0.92 & 10.53 & 1.35 & 6.77 & 1.68 \\
\texttt{+-{}+-{}-{}} & 5.89 & 1.04 & 6.6 & 1.28 & 10.43 & 1.45 & 6.89 & 1.9 \\
\texttt{+-{}-{}-{}-{}} & 5.37 & 0.9 & 5.69 & 1.31 & 10.35 & 1.31 & 7.08 & 1.96 \\
\hline
\end{tabular}

\caption{Creativity scores on the AUT for agents with different Big Five personality traits using Qwen2.5-32B. Scores include the four TTCT creativity assessment dimensions: ORIGINALITY, ELABORATION, FLUENCY, and FLEXIBILITY.}
\label{sin:aut:qwen32}
\end{table*}

\begin{table*}
\centering

\begin{tabular}{lcccccccc}
\hline
\multirow{2}{*}{Trait} & \multicolumn{2}{c}{ORIGINALITY} & \multicolumn{2}{c}{ELABORATION} & \multicolumn{2}{c}{FLUENCY} & \multicolumn{2}{c}{FLEXIBILITY} \\
\cline{2-9}
 & Mean & Std. & Mean & Std. & Mean & Std. & Mean & Std. \\
\hline
\texttt{++-{}-{}+} & 6.81 & 0.98 & 8.0 & 0.76 & 16.86 & 2.53 & 7.63 & 2.18 \\
\texttt{+++-{}+} & 6.75 & 0.94 & 7.99 & 0.77 & 17.33 & 2.64 & 7.66 & 1.96 \\
\texttt{-{}-{}+++} & 6.8 & 0.99 & 7.91 & 0.76 & 17.68 & 2.62 & 7.73 & 1.85 \\
\texttt{-{}-{}+-{}+} & 6.73 & 0.98 & 7.92 & 0.84 & 17.99 & 2.82 & 7.76 & 1.92 \\
\texttt{-{}+-{}-{}-{}} & 6.7 & 1.07 & 7.93 & 0.8 & 17.69 & 2.74 & 7.66 & 1.85 \\
\texttt{-{}+-{}-{}+} & 6.65 & 1.06 & 7.97 & 0.9 & 18.61 & 2.49 & 7.85 & 1.8 \\
\texttt{-{}++-{}+} & 6.72 & 0.98 & 7.9 & 0.77 & 18.62 & 2.28 & 7.71 & 1.76 \\
\texttt{-{}++++} & 6.74 & 1.03 & 7.88 & 0.82 & 18.6 & 2.62 & 7.84 & 1.92 \\
\texttt{++-{}++} & 6.75 & 1.07 & 7.86 & 0.71 & 17.53 & 2.96 & 7.89 & 2.12 \\
\texttt{+++++} & 6.65 & 1.14 & 7.89 & 0.82 & 18.16 & 2.72 & 7.37 & 1.74 \\
\texttt{+-{}-{}++} & 6.63 & 1.03 & 7.9 & 0.89 & 17.56 & 2.92 & 7.62 & 1.85 \\
\texttt{-{}++-{}-{}} & 6.56 & 0.99 & 7.9 & 0.77 & 18.15 & 2.9 & 7.65 & 1.77 \\
\texttt{-{}+++-{}} & 6.54 & 0.93 & 7.88 & 0.77 & 17.55 & 2.73 & 7.71 & 1.93 \\
\texttt{-{}+-{}+-{}} & 6.56 & 1.03 & 7.86 & 0.71 & 17.78 & 2.6 & 7.8 & 1.67 \\
\texttt{+-{}+-{}+} & 6.6 & 1.1 & 7.82 & 0.84 & 17.34 & 3.05 & 7.64 & 1.65 \\
\texttt{-{}-{}-{}-{}+} & 6.58 & 1.1 & 7.81 & 0.97 & 17.41 & 2.85 & 7.28 & 1.63 \\
\texttt{++++-{}} & 6.54 & 1.02 & 7.84 & 0.83 & 17.25 & 2.82 & 7.67 & 2.07 \\
\texttt{-{}+-{}++} & 6.52 & 1.06 & 7.86 & 0.85 & 18.34 & 2.79 & 7.5 & 2.28 \\
\texttt{-{}-{}++-{}} & 6.58 & 1.08 & 7.79 & 0.89 & 17.49 & 2.67 & 7.51 & 1.91 \\
\texttt{+-{}+++} & 6.53 & 1.0 & 7.78 & 0.86 & 17.74 & 2.81 & 7.57 & 1.42 \\
\texttt{+++-{}-{}} & 6.46 & 0.98 & 7.85 & 0.82 & 16.9 & 2.59 & 7.3 & 2.19 \\
\texttt{+-{}-{}-{}+} & 6.55 & 1.24 & 7.76 & 1.05 & 16.88 & 2.82 & 7.44 & 1.86 \\
\texttt{-{}-{}-{}-{}-{}} & 6.6 & 1.04 & 7.69 & 0.82 & 17.65 & 2.7 & 7.38 & 1.92 \\
\texttt{-{}-{}-{}++} & 6.53 & 1.14 & 7.75 & 0.79 & 17.88 & 2.85 & 7.46 & 1.99 \\
\texttt{-{}-{}+-{}-{}} & 6.48 & 1.19 & 7.78 & 1.0 & 18.16 & 2.74 & 7.29 & 1.66 \\
\texttt{-{}-{}-{}+-{}} & 6.51 & 1.11 & 7.71 & 0.9 & 17.36 & 3.1 & 7.4 & 1.88 \\
\texttt{+-{}+-{}-{}} & 6.37 & 1.15 & 7.69 & 1.11 & 16.27 & 3.35 & 7.18 & 1.65 \\
\texttt{++-{}-{}-{}} & 6.44 & 1.03 & 7.6 & 0.99 & 16.7 & 3.17 & 7.53 & 1.79 \\
\texttt{++-{}+-{}} & 6.33 & 1.05 & 7.65 & 1.05 & 17.79 & 2.45 & 7.64 & 2.13 \\
\texttt{+-{}++-{}} & 6.29 & 1.06 & 7.65 & 0.84 & 16.99 & 3.42 & 7.71 & 2.21 \\
\texttt{+-{}-{}-{}-{}} & 6.33 & 1.18 & 7.57 & 1.09 & 16.58 & 3.2 & 7.48 & 2.18 \\
\texttt{+-{}-{}+-{}} & 6.27 & 1.13 & 7.53 & 1.05 & 16.94 & 3.17 & 7.39 & 1.85 \\
\hline
\end{tabular}

\caption{Creativity scores on the AUT for agents with different Big Five personality traits using Qwen2.5-14B. Scores include the four TTCT creativity assessment dimensions: ORIGINALITY, ELABORATION, FLUENCY, and FLEXIBILITY.}
\label{sin:aut:14B}
\end{table*}

\begin{table*}
\centering

\begin{tabular}{lcccccccc}
\hline
\multirow{2}{*}{Trait} & \multicolumn{2}{c}{ORIGINALITY} & \multicolumn{2}{c}{ELABORATION} & \multicolumn{2}{c}{FLUENCY} & \multicolumn{2}{c}{FLEXIBILITY} \\
\cline{2-9}
 & Mean & Std. & Mean & Std. & Mean & Std. & Mean & Std. \\
\hline
\texttt{-{}-{}+++} & 6.12 & 1.0 & 7.79 & 0.84 & 19.75 & 2.25 & 7.11 & 1.39 \\
\texttt{-{}+-{}-{}+} & 6.13 & 1.12 & 7.75 & 0.84 & 20.29 & 3.52 & 6.39 & 1.74 \\
\texttt{-{}+-{}++} & 6.17 & 0.93 & 7.69 & 0.73 & 20.24 & 3.27 & 5.8 & 1.57 \\
\texttt{-{}++-{}+} & 6.13 & 0.95 & 7.69 & 0.73 & 20.15 & 2.6 & 6.3 & 1.75 \\
\texttt{-{}++++} & 5.96 & 0.87 & 7.75 & 0.64 & 20.89 & 3.47 & 5.99 & 1.45 \\
\texttt{-{}+-{}+-{}} & 5.95 & 0.96 & 7.6 & 0.85 & 20.2 & 2.55 & 6.73 & 1.73 \\
\texttt{-{}+++-{}} & 5.48 & 0.82 & 7.34 & 0.82 & 20.76 & 2.61 & 6.06 & 1.66 \\
\texttt{-{}++-{}-{}} & 5.58 & 0.93 & 7.24 & 0.87 & 18.34 & 4.45 & 6.99 & 1.51 \\
\texttt{-{}+-{}-{}-{}} & 5.72 & 0.95 & 7.05 & 0.99 & 18.32 & 5.8 & 6.88 & 1.7 \\
\texttt{++-{}++} & 5.58 & 1.12 & 6.65 & 1.29 & 15.88 & 2.92 & 6.92 & 1.78 \\
\texttt{+++++} & 5.48 & 1.02 & 6.71 & 1.22 & 16.28 & 3.18 & 7.02 & 1.62 \\
\texttt{++++-{}} & 5.41 & 0.86 & 6.63 & 1.25 & 15.44 & 2.83 & 6.98 & 1.9 \\
\texttt{+++-{}+} & 5.33 & 1.04 & 6.54 & 1.3 & 15.59 & 2.62 & 6.85 & 1.95 \\
\texttt{++-{}-{}+} & 5.37 & 1.0 & 6.39 & 1.36 & 15.26 & 2.61 & 6.72 & 1.5 \\
\texttt{+++-{}-{}} & 5.25 & 0.93 & 6.36 & 1.28 & 14.6 & 2.66 & 7.11 & 1.86 \\
\texttt{++-{}+-{}} & 5.27 & 0.9 & 6.32 & 1.31 & 15.62 & 2.41 & 7.05 & 1.97 \\
\texttt{++-{}-{}-{}} & 5.18 & 0.99 & 6.1 & 1.34 & 14.67 & 2.27 & 6.79 & 1.74 \\
\texttt{-{}-{}++-{}} & 4.97 & 1.01 & 6.07 & 1.5 & 16.12 & 2.78 & 7.07 & 1.55 \\
\texttt{-{}-{}+-{}+} & 5.03 & 1.15 & 5.94 & 1.57 & 16.25 & 2.63 & 6.48 & 1.62 \\
\texttt{-{}-{}+-{}-{}} & 4.96 & 1.06 & 5.92 & 1.45 & 14.99 & 2.78 & 6.66 & 1.83 \\
\texttt{-{}-{}-{}+-{}} & 4.91 & 0.85 & 5.85 & 1.42 & 14.9 & 2.49 & 7.09 & 1.93 \\
\texttt{-{}-{}-{}-{}+} & 4.92 & 0.97 & 5.72 & 1.38 & 15.71 & 2.55 & 6.92 & 1.81 \\
\texttt{-{}-{}-{}-{}-{}} & 4.92 & 0.95 & 5.7 & 1.44 & 15.29 & 2.48 & 6.87 & 1.51 \\
\texttt{-{}-{}-{}++} & 4.8 & 1.07 & 5.44 & 1.4 & 14.96 & 2.69 & 6.78 & 1.57 \\
\texttt{+-{}+-{}-{}} & 4.33 & 0.98 & 4.82 & 1.09 & 14.63 & 2.08 & 6.49 & 1.51 \\
\texttt{+-{}++-{}} & 4.28 & 0.86 & 4.82 & 0.96 & 14.85 & 2.43 & 6.9 & 1.72 \\
\texttt{+-{}+-{}+} & 4.28 & 0.75 & 4.81 & 0.99 & 15.21 & 2.52 & 6.58 & 1.48 \\
\texttt{+-{}+++} & 4.17 & 1.05 & 4.78 & 1.04 & 14.81 & 2.42 & 6.79 & 1.8 \\
\texttt{+-{}-{}-{}-{}} & 4.23 & 0.79 & 4.59 & 0.94 & 14.53 & 1.72 & 6.81 & 2.07 \\
\texttt{+-{}-{}+-{}} & 4.15 & 1.01 & 4.57 & 1.02 & 14.89 & 2.15 & 6.71 & 1.72 \\
\texttt{+-{}-{}-{}+} & 4.12 & 0.89 & 4.57 & 0.76 & 15.36 & 2.1 & 6.53 & 1.37 \\
\texttt{+-{}-{}++} & 4.04 & 0.92 & 4.64 & 1.1 & 15.67 & 2.52 & 6.67 & 1.42 \\
\hline
\end{tabular}

\caption{Creativity scores on the AUT for agents with different Big Five personality traits using Llama3.1-8B. Scores include the four TTCT creativity assessment dimensions: ORIGINALITY, ELABORATION, FLUENCY, and FLEXIBILITY.}
\label{sin:aut:llama}
\end{table*}

\label{s:create}
In Table~\ref{sin:aut:qwen32}-\ref{sin:SIMILARITIES:llama}, we present the creativity performance of agents with different Big Five personality traits in various open tasks. The results show that agents with different personality traits exhibit significant differences in creativity levels.

\begin{table*}
\centering

\begin{tabular}{lcccccccc}
\hline
\multirow{2}{*}{Trait} & \multicolumn{2}{c}{ORIGINALITY} & \multicolumn{2}{c}{ELABORATION} & \multicolumn{2}{c}{FLUENCY} & \multicolumn{2}{c}{FLEXIBILITY} \\
\cline{2-9}
 & Mean & Std. & Mean & Std. & Mean & Std. & Mean & Std. \\
\hline
\texttt{+++-{}+} & 3.76 & 0.97 & 6.88 & 1.84 & 16.33 & 10.18 & 7.68 & 2.33 \\
\texttt{++-{}++} & 3.7 & 0.87 & 6.9 & 1.68 & 12.22 & 6.11 & 7.32 & 2.11 \\
\texttt{+++++} & 3.63 & 0.91 & 6.88 & 1.86 & 17.84 & 12.47 & 7.98 & 3.0 \\
\texttt{-{}+-{}-{}+} & 3.82 & 1.02 & 6.41 & 2.04 & 14.6 & 8.4 & 7.44 & 2.37 \\
\texttt{-{}++++} & 3.8 & 0.95 & 6.32 & 2.04 & 20.53 & 13.52 & 8.21 & 2.98 \\
\texttt{++-{}-{}+} & 3.61 & 0.95 & 6.5 & 1.74 & 12.09 & 4.99 & 6.92 & 2.39 \\
\texttt{-{}++-{}+} & 3.68 & 0.93 & 6.12 & 2.04 & 18.02 & 10.36 & 8.18 & 2.89 \\
\texttt{+-{}+++} & 3.73 & 0.97 & 6.06 & 2.18 & 12.28 & 5.3 & 6.72 & 2.41 \\
\texttt{+-{}-{}++} & 3.82 & 0.95 & 5.91 & 2.21 & 12.27 & 5.23 & 6.81 & 2.18 \\
\texttt{-{}+-{}++} & 3.36 & 1.05 & 5.86 & 2.03 & 17.76 & 11.61 & 8.06 & 3.09 \\
\texttt{++-{}+-{}} & 3.28 & 0.92 & 5.68 & 1.89 & 11.46 & 4.82 & 6.62 & 2.44 \\
\texttt{-{}-{}+++} & 3.34 & 0.97 & 5.53 & 2.29 & 16.43 & 9.51 & 7.53 & 2.46 \\
\texttt{-{}+-{}+-{}} & 3.41 & 0.99 & 5.35 & 2.13 & 13.45 & 5.8 & 7.29 & 2.55 \\
\texttt{++++-{}} & 3.35 & 0.85 & 5.36 & 2.26 & 14.52 & 8.19 & 7.02 & 2.08 \\
\texttt{-{}+++-{}} & 3.31 & 0.88 & 5.26 & 2.02 & 16.91 & 10.47 & 7.41 & 2.67 \\
\texttt{-{}-{}+-{}+} & 3.37 & 0.98 & 4.84 & 2.12 & 16.5 & 9.05 & 7.27 & 2.43 \\
\texttt{+-{}+-{}+} & 3.3 & 1.0 & 4.88 & 2.21 & 13.13 & 5.69 & 7.06 & 2.61 \\
\texttt{-{}++-{}-{}} & 3.17 & 0.88 & 4.91 & 2.0 & 14.19 & 6.83 & 7.48 & 2.83 \\
\texttt{+++-{}-{}} & 3.19 & 0.99 & 4.88 & 2.26 & 13.82 & 9.44 & 6.91 & 2.43 \\
\texttt{-{}-{}-{}++} & 3.37 & 0.95 & 4.64 & 2.06 & 13.73 & 4.61 & 7.1 & 2.53 \\
\texttt{+-{}-{}-{}+} & 3.37 & 1.1 & 4.27 & 2.37 & 11.03 & 3.73 & 6.66 & 2.21 \\
\texttt{-{}-{}++-{}} & 3.28 & 0.95 & 4.12 & 1.97 & 13.59 & 5.6 & 6.86 & 2.18 \\
\texttt{-{}-{}-{}-{}+} & 3.15 & 1.08 & 4.23 & 2.14 & 13.87 & 5.75 & 6.71 & 2.41 \\
\texttt{++-{}-{}-{}} & 2.96 & 0.89 & 4.33 & 1.89 & 11.44 & 5.12 & 6.49 & 2.26 \\
\texttt{-{}-{}-{}+-{}} & 3.18 & 0.96 & 3.86 & 1.85 & 13.23 & 5.65 & 6.59 & 2.36 \\
\texttt{-{}+-{}-{}-{}} & 2.87 & 0.89 & 4.0 & 1.78 & 12.88 & 6.64 & 6.7 & 2.14 \\
\texttt{+-{}-{}+-{}} & 3.2 & 0.93 & 3.67 & 1.59 & 10.55 & 3.84 & 5.73 & 1.78 \\
\texttt{+-{}++-{}} & 3.12 & 0.97 & 3.64 & 1.93 & 11.64 & 3.64 & 6.33 & 1.91 \\
\texttt{-{}-{}+-{}-{}} & 2.78 & 0.86 & 3.39 & 1.54 & 12.72 & 4.65 & 6.67 & 2.49 \\
\texttt{+-{}+-{}-{}} & 2.79 & 0.9 & 3.12 & 1.7 & 11.25 & 4.65 & 5.99 & 1.91 \\
\texttt{-{}-{}-{}-{}-{}} & 2.55 & 0.75 & 3.05 & 1.63 & 12.22 & 5.94 & 6.13 & 2.08 \\
\texttt{+-{}-{}-{}-{}} & 2.6 & 0.87 & 2.49 & 1.16 & 9.08 & 3.61 & 5.31 & 1.85 \\
\hline
\end{tabular}

\caption{Creativity scores on INSTANCES for agents with different Big Five personality traits using Qwen2.5-32B. Scores include the four TTCT creativity assessment dimensions: ORIGINALITY, ELABORATION, FLUENCY, and FLEXIBILITY.}
\label{INSTANCES:qwen32}
\end{table*}

       \begin{table*}
\centering

\begin{tabular}{lcccccccc}
\hline
\multirow{2}{*}{Trait} & \multicolumn{2}{c}{ORIGINALITY} & \multicolumn{2}{c}{ELABORATION} & \multicolumn{2}{c}{FLUENCY} & \multicolumn{2}{c}{FLEXIBILITY} \\
\cline{2-9}
 & Mean & Std. & Mean & Std. & Mean & Std. & Mean & Std. \\
\hline
\texttt{-{}++-{}+} & 3.41 & 0.97 & 5.03 & 1.94 & 18.65 & 12.88 & 7.97 & 2.89 \\
\texttt{+++-{}+} & 3.34 & 0.97 & 5.06 & 1.88 & 14.57 & 9.31 & 7.46 & 2.28 \\
\texttt{+++++} & 3.4 & 1.0 & 4.98 & 1.79 & 17.16 & 12.76 & 7.68 & 2.83 \\
\texttt{-{}+-{}-{}+} & 3.4 & 0.95 & 4.94 & 1.75 & 16.6 & 11.05 & 8.08 & 3.01 \\
\texttt{-{}++++} & 3.52 & 0.92 & 4.82 & 1.75 & 18.45 & 11.58 & 8.39 & 3.49 \\
\texttt{-{}+-{}++} & 3.47 & 0.94 & 4.69 & 1.68 & 17.12 & 11.98 & 8.16 & 3.28 \\
\texttt{++-{}-{}+} & 3.34 & 1.0 & 4.78 & 1.81 & 11.91 & 5.14 & 6.93 & 2.17 \\
\texttt{++-{}++} & 3.35 & 0.94 & 4.74 & 1.75 & 13.61 & 6.53 & 7.37 & 2.72 \\
\texttt{-{}++-{}-{}} & 3.11 & 0.88 & 4.59 & 1.79 & 14.02 & 7.1 & 7.3 & 2.65 \\
\texttt{-{}+++-{}} & 3.16 & 0.94 & 4.48 & 1.63 & 14.15 & 7.24 & 7.4 & 2.59 \\
\texttt{-{}+-{}-{}-{}} & 3.27 & 0.93 & 4.31 & 1.7 & 13.97 & 7.6 & 7.35 & 2.65 \\
\texttt{-{}-{}+++} & 3.25 & 0.88 & 4.31 & 1.6 & 14.61 & 7.75 & 7.23 & 2.8 \\
\texttt{-{}+-{}+-{}} & 3.1 & 0.96 & 4.41 & 1.66 & 13.76 & 7.21 & 7.68 & 2.77 \\
\texttt{+-{}+++} & 3.24 & 1.01 & 4.25 & 1.66 & 12.49 & 6.05 & 6.8 & 2.43 \\
\texttt{++-{}+-{}} & 3.06 & 0.93 & 4.27 & 1.62 & 11.49 & 6.17 & 6.76 & 2.25 \\
\texttt{-{}-{}+-{}+} & 3.1 & 0.98 & 4.22 & 1.71 & 14.2 & 7.93 & 6.74 & 2.36 \\
\texttt{++++-{}} & 3.11 & 0.92 & 4.2 & 1.59 & 12.33 & 6.51 & 6.68 & 2.38 \\
\texttt{++-{}-{}-{}} & 3.0 & 0.97 & 4.23 & 1.67 & 10.76 & 5.59 & 6.22 & 2.18 \\
\texttt{+++-{}-{}} & 2.96 & 0.86 & 4.22 & 1.66 & 10.93 & 4.55 & 6.43 & 2.26 \\
\texttt{-{}-{}+-{}-{}} & 3.11 & 0.89 & 3.88 & 1.54 & 13.25 & 7.16 & 6.82 & 2.35 \\
\texttt{-{}-{}++-{}} & 3.02 & 0.85 & 3.93 & 1.55 & 12.57 & 5.66 & 6.87 & 2.44 \\
\texttt{-{}-{}-{}++} & 3.0 & 0.92 & 3.86 & 1.52 & 13.44 & 7.65 & 6.9 & 2.44 \\
\texttt{+-{}-{}++} & 3.11 & 1.1 & 3.74 & 1.71 & 10.34 & 4.6 & 6.29 & 2.31 \\
\texttt{-{}-{}-{}+-{}} & 2.96 & 0.87 & 3.73 & 1.33 & 12.63 & 5.88 & 6.65 & 2.41 \\
\texttt{-{}-{}-{}-{}+} & 3.07 & 0.87 & 3.57 & 1.49 & 12.79 & 7.0 & 6.68 & 2.21 \\
\texttt{-{}-{}-{}-{}-{}} & 2.93 & 0.94 & 3.35 & 1.38 & 11.31 & 5.21 & 6.3 & 2.15 \\
\texttt{+-{}+-{}+} & 2.87 & 0.96 & 3.09 & 1.46 & 9.23 & 2.45 & 5.55 & 2.02 \\
\texttt{+-{}++-{}} & 2.6 & 0.85 & 3.19 & 1.25 & 8.87 & 3.19 & 5.5 & 2.18 \\
\texttt{+-{}-{}+-{}} & 2.46 & 0.71 & 2.81 & 1.23 & 7.95 & 3.47 & 4.78 & 2.29 \\
\texttt{+-{}-{}-{}+} & 2.42 & 0.91 & 2.48 & 1.31 & 7.32 & 4.53 & 4.8 & 2.52 \\
\texttt{+-{}+-{}-{}} & 2.34 & 0.78 & 2.5 & 1.35 & 6.96 & 3.11 & 4.48 & 1.78 \\
\texttt{+-{}-{}-{}-{}} & 2.04 & 0.62 & 2.04 & 0.93 & 5.73 & 2.58 & 3.67 & 1.41 \\
\hline
\end{tabular}

\caption{Creativity scores on INSTANCES for agents with different Big Five personality traits using Qwen2.5-14B. Scores include the four TTCT creativity assessment dimensions: ORIGINALITY, ELABORATION, FLUENCY, and FLEXIBILITY.}
\label{INSTANCE:qwen14}
\end{table*}

\begin{table*}
\centering

\begin{tabular}{lcccccccc}
\hline
\multirow{2}{*}{Trait} & \multicolumn{2}{c}{ORIGINALITY} & \multicolumn{2}{c}{ELABORATION} & \multicolumn{2}{c}{FLUENCY} & \multicolumn{2}{c}{FLEXIBILITY} \\
\cline{2-9}
 & Mean & Std. & Mean & Std. & Mean & Std. & Mean & Std. \\
\hline
\texttt{-{}+-{}++} & 3.8 & 0.95 & 6.12 & 1.84 & 24.7 & 17.72 & 9.24 & 4.08 \\
\texttt{-{}++-{}+} & 3.83 & 1.04 & 5.84 & 1.78 & 33.16 & 25.99 & 8.42 & 3.93 \\
\texttt{-{}+-{}-{}+} & 3.66 & 0.98 & 5.67 & 1.81 & 26.78 & 20.79 & 7.97 & 3.74 \\
\texttt{-{}+++-{}} & 3.47 & 0.96 & 4.89 & 1.56 & 29.04 & 20.15 & 7.28 & 2.72 \\
\texttt{-{}+-{}+-{}} & 3.38 & 0.95 & 4.84 & 1.78 & 23.89 & 17.69 & 7.57 & 3.45 \\
\texttt{+++++} & 3.38 & 1.01 & 4.51 & 1.38 & 30.59 & 25.32 & 7.93 & 2.93 \\
\texttt{++-{}++} & 3.52 & 0.97 & 4.08 & 1.47 & 28.43 & 21.59 & 7.39 & 3.04 \\
\texttt{+++-{}+} & 3.32 & 0.88 & 4.26 & 1.38 & 29.58 & 21.0 & 7.57 & 2.76 \\
\texttt{-{}-{}+++} & 3.4 & 0.98 & 4.09 & 1.86 & 23.73 & 14.13 & 8.42 & 3.97 \\
\texttt{-{}++-{}-{}} & 3.06 & 0.91 & 3.92 & 1.54 & 27.96 & 24.26 & 7.32 & 3.19 \\
\texttt{+++-{}-{}} & 3.21 & 0.87 & 3.74 & 1.15 & 31.34 & 21.62 & 7.14 & 3.04 \\
\texttt{++++-{}} & 3.18 & 0.92 & 3.76 & 1.24 & 29.59 & 22.51 & 7.64 & 2.9 \\
\texttt{++-{}-{}+} & 3.17 & 0.91 & 3.68 & 1.19 & 28.86 & 21.48 & 7.43 & 2.93 \\
\texttt{-{}+-{}-{}-{}} & 3.07 & 0.93 & 3.61 & 1.29 & 24.35 & 22.33 & 6.8 & 2.95 \\
\texttt{-{}-{}+-{}+} & 2.97 & 0.94 & 3.54 & 1.49 & 25.71 & 20.62 & 7.58 & 3.14 \\
\texttt{++-{}+-{}} & 2.96 & 0.93 & 3.45 & 1.03 & 26.7 & 20.69 & 7.17 & 3.18 \\
\texttt{+-{}+++} & 2.92 & 0.82 & 3.19 & 1.29 & 24.55 & 19.09 & 7.46 & 2.86 \\
\texttt{-{}-{}-{}-{}+} & 2.87 & 0.81 & 3.11 & 1.22 & 25.62 & 19.0 & 7.47 & 2.75 \\
\texttt{+-{}-{}++} & 2.94 & 1.05 & 3.0 & 1.15 & 27.2 & 22.5 & 7.94 & 2.87 \\
\texttt{+-{}+-{}+} & 2.94 & 0.96 & 2.91 & 1.21 & 25.62 & 18.58 & 7.57 & 3.11 \\
\texttt{-{}-{}-{}++} & 2.79 & 0.9 & 3.04 & 1.12 & 26.7 & 21.79 & 7.56 & 3.14 \\
\texttt{-{}-{}+-{}-{}} & 2.78 & 0.77 & 2.96 & 1.22 & 26.42 & 21.79 & 7.31 & 3.04 \\
\texttt{-{}-{}++-{}} & 2.93 & 0.84 & 2.71 & 1.07 & 24.03 & 20.47 & 7.32 & 2.62 \\
\texttt{+-{}-{}-{}+} & 2.84 & 0.91 & 2.78 & 1.03 & 23.76 & 15.28 & 7.59 & 2.58 \\
\texttt{-{}-{}-{}-{}-{}} & 2.76 & 0.85 & 2.83 & 0.95 & 30.63 & 25.18 & 7.3 & 3.1 \\
\texttt{+-{}++-{}} & 2.8 & 0.87 & 2.69 & 0.96 & 26.61 & 18.93 & 7.65 & 2.92 \\
\texttt{+-{}-{}+-{}} & 2.74 & 0.89 & 2.6 & 0.93 & 23.81 & 15.5 & 6.92 & 2.25 \\
\texttt{+-{}+-{}-{}} & 2.7 & 0.88 & 2.61 & 1.01 & 31.3 & 27.17 & 7.49 & 2.63 \\
\texttt{-{}-{}-{}+-{}} & 2.54 & 0.8 & 2.72 & 0.92 & 25.71 & 22.37 & 6.9 & 2.58 \\
\texttt{+-{}-{}-{}-{}} & 2.67 & 0.84 & 2.46 & 0.99 & 24.78 & 20.6 & 7.0 & 2.71 \\
\hline
\end{tabular}

\caption{Creativity scores on the INSTANCES for agents with different Big Five personality traits using Llama3.1-8B. Scores include the four TTCT creativity assessment dimensions: ORIGINALITY, ELABORATION, FLUENCY, and FLEXIBILITY.}
\label{sin:INS:Llama}
\end{table*}

   \begin{table*}
\centering

\begin{tabular}{lcccccccc}
\hline
\multirow{2}{*}{Trait} & \multicolumn{2}{c}{ORIGINALITY} & \multicolumn{2}{c}{ELABORATION} & \multicolumn{2}{c}{FLUENCY} & \multicolumn{2}{c}{FLEXIBILITY} \\
\cline{2-9}
 & Mean & Std. & Mean & Std. & Mean & Std. & Mean & Std. \\
\hline
\texttt{+++++} & 4.91 & 1.0 & 7.68 & 0.88 & 6.78 & 2.31 & 5.71 & 1.24 \\
\texttt{-{}++++} & 4.76 & 1.02 & 7.67 & 1.09 & 6.94 & 2.61 & 5.79 & 1.29 \\
\texttt{-{}+-{}-{}+} & 4.82 & 0.94 & 7.61 & 0.89 & 7.0 & 3.48 & 5.46 & 1.14 \\
\texttt{+++-{}+} & 4.74 & 0.85 & 7.68 & 0.88 & 6.81 & 2.27 & 5.65 & 1.24 \\
\texttt{-{}+-{}++} & 4.76 & 0.98 & 7.66 & 0.93 & 6.64 & 2.68 & 5.7 & 1.35 \\
\texttt{-{}++-{}+} & 4.6 & 0.87 & 7.8 & 0.93 & 6.73 & 2.33 & 5.69 & 1.49 \\
\texttt{-{}-{}+++} & 4.88 & 0.97 & 7.51 & 0.97 & 6.48 & 2.24 & 5.51 & 1.27 \\
\texttt{++-{}++} & 4.97 & 1.0 & 7.39 & 1.02 & 6.65 & 2.65 & 5.47 & 1.18 \\
\texttt{++-{}-{}+} & 4.9 & 1.05 & 7.38 & 1.01 & 6.75 & 2.67 & 5.33 & 1.06 \\
\texttt{+-{}+++} & 4.94 & 0.98 & 7.3 & 1.19 & 6.57 & 2.37 & 5.5 & 1.12 \\
\texttt{-{}+++-{}} & 4.54 & 0.79 & 7.57 & 0.91 & 6.71 & 2.43 & 5.75 & 1.34 \\
\texttt{-{}-{}+-{}+} & 4.64 & 1.01 & 7.41 & 0.97 & 6.82 & 2.4 & 5.59 & 1.4 \\
\texttt{-{}++-{}-{}} & 4.58 & 0.9 & 7.39 & 1.08 & 6.7 & 2.54 & 5.51 & 1.41 \\
\texttt{+-{}+-{}+} & 4.8 & 0.97 & 7.16 & 1.15 & 6.7 & 2.93 & 5.4 & 1.05 \\
\texttt{+-{}-{}++} & 5.05 & 1.17 & 6.85 & 1.29 & 6.89 & 2.56 & 5.48 & 1.17 \\
\texttt{-{}-{}-{}-{}+} & 4.73 & 0.92 & 7.14 & 1.09 & 6.56 & 3.15 & 5.25 & 1.18 \\
\texttt{-{}+-{}+-{}} & 4.66 & 0.87 & 7.2 & 1.06 & 6.69 & 2.77 & 5.5 & 1.26 \\
\texttt{++++-{}} & 4.57 & 0.84 & 7.17 & 1.21 & 6.9 & 3.72 & 5.51 & 1.11 \\
\texttt{-{}-{}-{}++} & 4.6 & 0.95 & 7.09 & 1.16 & 6.24 & 2.63 & 5.32 & 1.2 \\
\texttt{-{}-{}++-{}} & 4.56 & 0.93 & 6.99 & 1.16 & 6.46 & 2.51 & 5.34 & 1.26 \\
\texttt{+++-{}-{}} & 4.55 & 0.89 & 6.99 & 1.14 & 6.55 & 1.98 & 5.49 & 1.22 \\
\texttt{+-{}-{}-{}+} & 4.87 & 1.13 & 6.59 & 1.41 & 6.84 & 2.56 & 5.21 & 1.12 \\
\texttt{-{}+-{}-{}-{}} & 4.32 & 0.83 & 7.03 & 1.08 & 6.85 & 2.61 & 5.3 & 1.61 \\
\texttt{-{}-{}+-{}-{}} & 4.37 & 0.86 & 6.87 & 1.25 & 6.35 & 2.47 & 5.17 & 1.3 \\
\texttt{++-{}-{}-{}} & 4.42 & 0.89 & 6.49 & 1.24 & 6.91 & 2.97 & 5.21 & 1.27 \\
\texttt{++-{}+-{}} & 4.49 & 0.9 & 6.4 & 1.19 & 6.95 & 2.84 & 5.32 & 1.08 \\
\texttt{+-{}++-{}} & 4.49 & 1.02 & 6.3 & 1.66 & 6.88 & 3.12 & 5.17 & 1.01 \\
\texttt{+-{}+-{}-{}} & 4.37 & 0.98 & 6.09 & 1.48 & 6.8 & 2.63 & 4.98 & 1.16 \\
\texttt{-{}-{}-{}+-{}} & 4.34 & 0.92 & 5.98 & 1.57 & 6.67 & 2.41 & 5.0 & 1.1 \\
\texttt{+-{}-{}+-{}} & 4.47 & 1.06 & 5.38 & 1.64 & 6.81 & 2.69 & 5.05 & 1.09 \\
\texttt{-{}-{}-{}-{}-{}} & 4.07 & 1.0 & 5.66 & 1.54 & 6.88 & 2.37 & 4.88 & 1.02 \\
\texttt{+-{}-{}-{}-{}} & 3.55 & 1.11 & 4.14 & 1.55 & 6.91 & 2.56 & 4.66 & 1.0 \\
\hline
\end{tabular}

\caption{Creativity scores on Scientific for agents with different Big Five personality traits using Qwen2.5-32B. Scores include the four TTCT creativity assessment dimensions: ORIGINALITY, ELABORATION, FLUENCY, and FLEXIBILITY.}
\label{sin:sci:32B}
\end{table*}

\begin{table*}
\centering

\begin{tabular}{lcccccccc}
\hline
\multirow{2}{*}{Trait} & \multicolumn{2}{c}{ORIGINALITY} & \multicolumn{2}{c}{ELABORATION} & \multicolumn{2}{c}{FLUENCY} & \multicolumn{2}{c}{FLEXIBILITY} \\
\cline{2-9}
 & Mean & Std. & Mean & Std. & Mean & Std. & Mean & Std. \\
\hline
\texttt{-{}++-{}+} & 4.85 & 1.09 & 7.9 & 0.8 & 7.17 & 2.67 & 5.84 & 1.51 \\
\texttt{-{}++++} & 4.83 & 1.06 & 7.9 & 0.72 & 7.23 & 2.39 & 5.96 & 1.41 \\
\texttt{+++-{}+} & 4.96 & 1.18 & 7.75 & 0.96 & 7.21 & 2.88 & 5.64 & 1.28 \\
\texttt{-{}+-{}-{}+} & 4.72 & 1.05 & 7.9 & 0.78 & 7.19 & 2.27 & 5.95 & 1.42 \\
\texttt{+++++} & 4.87 & 1.0 & 7.7 & 0.87 & 6.99 & 2.61 & 5.68 & 1.21 \\
\texttt{-{}+-{}++} & 4.78 & 1.03 & 7.78 & 0.87 & 7.33 & 2.56 & 5.87 & 1.41 \\
\texttt{-{}++-{}-{}} & 4.71 & 0.99 & 7.66 & 0.87 & 6.83 & 2.47 & 5.67 & 1.36 \\
\texttt{-{}-{}+++} & 4.69 & 1.02 & 7.65 & 0.88 & 7.0 & 2.47 & 5.81 & 1.47 \\
\texttt{-{}+-{}+-{}} & 4.7 & 1.18 & 7.6 & 1.01 & 7.02 & 2.76 & 5.67 & 1.49 \\
\texttt{++-{}-{}+} & 4.85 & 0.98 & 7.44 & 1.03 & 7.44 & 2.82 & 5.64 & 1.36 \\
\texttt{++-{}++} & 4.74 & 1.15 & 7.48 & 1.04 & 7.31 & 3.07 & 5.57 & 1.31 \\
\texttt{-{}-{}+-{}+} & 4.66 & 1.0 & 7.55 & 0.96 & 7.14 & 2.56 & 5.63 & 1.33 \\
\texttt{-{}+-{}-{}-{}} & 4.71 & 1.0 & 7.5 & 0.99 & 7.22 & 2.64 & 5.47 & 1.45 \\
\texttt{-{}+++-{}} & 4.55 & 0.93 & 7.56 & 0.9 & 7.05 & 2.58 & 5.7 & 1.47 \\
\texttt{+-{}+++} & 4.78 & 1.1 & 7.23 & 1.14 & 7.5 & 4.09 & 5.61 & 1.23 \\
\texttt{-{}-{}-{}-{}+} & 4.55 & 1.04 & 7.27 & 1.02 & 6.89 & 2.37 & 5.46 & 1.45 \\
\texttt{-{}-{}+-{}-{}} & 4.44 & 0.86 & 7.31 & 1.04 & 6.99 & 2.55 & 5.35 & 1.32 \\
\texttt{-{}-{}++-{}} & 4.5 & 1.02 & 7.24 & 1.2 & 6.66 & 2.86 & 5.34 & 1.37 \\
\texttt{-{}-{}-{}++} & 4.51 & 0.85 & 7.2 & 1.16 & 7.12 & 3.23 & 5.5 & 1.47 \\
\texttt{+++-{}-{}} & 4.57 & 0.93 & 6.99 & 1.31 & 7.46 & 2.89 & 5.47 & 1.49 \\
\texttt{++++-{}} & 4.56 & 1.01 & 6.94 & 1.23 & 7.11 & 2.27 & 5.26 & 1.36 \\
\texttt{+-{}+-{}+} & 4.55 & 0.98 & 6.57 & 1.44 & 7.68 & 2.91 & 5.43 & 1.14 \\
\texttt{-{}-{}-{}+-{}} & 4.45 & 0.93 & 6.58 & 1.21 & 7.08 & 3.14 & 5.21 & 1.26 \\
\texttt{-{}-{}-{}-{}-{}} & 4.27 & 1.02 & 6.7 & 1.4 & 7.47 & 3.95 & 5.09 & 1.2 \\
\texttt{++-{}-{}-{}} & 4.49 & 0.88 & 6.42 & 1.26 & 7.5 & 2.86 & 5.15 & 1.26 \\
\texttt{++-{}+-{}} & 4.39 & 0.85 & 6.44 & 1.21 & 7.53 & 2.51 & 5.42 & 1.29 \\
\texttt{+-{}-{}++} & 4.44 & 1.24 & 5.9 & 1.81 & 7.8 & 3.41 & 5.26 & 1.2 \\
\texttt{+-{}-{}-{}+} & 4.11 & 1.46 & 5.31 & 2.03 & 7.86 & 3.09 & 5.11 & 1.22 \\
\texttt{+-{}++-{}} & 3.83 & 1.18 & 4.85 & 1.72 & 7.62 & 3.12 & 4.86 & 1.13 \\
\texttt{+-{}+-{}-{}} & 3.49 & 1.32 & 4.72 & 2.06 & 7.9 & 2.53 & 4.54 & 1.23 \\
\texttt{+-{}-{}+-{}} & 3.47 & 1.2 & 4.06 & 1.88 & 8.11 & 3.03 & 4.76 & 1.15 \\
\texttt{+-{}-{}-{}-{}} & 3.14 & 1.48 & 3.75 & 2.07 & 7.74 & 2.9 & 4.54 & 1.21 \\
\hline
\end{tabular}

\caption{Creativity scores on SCIENTIFIC for agents with different Big Five personality traits using Qwen2.5-14B. Scores include the four TTCT creativity assessment dimensions: ORIGINALITY, ELABORATION, FLUENCY, and FLEXIBILITY.}
\label{sin:sci:14B}
\end{table*}

\begin{table*}
\centering

\begin{tabular}{lcccccccc}
\hline
\multirow{2}{*}{Trait} & \multicolumn{2}{c}{ORIGINALITY} & \multicolumn{2}{c}{ELABORATION} & \multicolumn{2}{c}{FLUENCY} & \multicolumn{2}{c}{FLEXIBILITY} \\
\cline{2-9}
 & Mean & Std. & Mean & Std. & Mean & Std. & Mean & Std. \\
\hline
\texttt{-{}++-{}+} & 4.85 & 1.09 & 7.9 & 0.8 & 7.17 & 2.67 & 5.84 & 1.51 \\
\texttt{-{}++++} & 4.83 & 1.06 & 7.9 & 0.72 & 7.23 & 2.39 & 5.96 & 1.41 \\
\texttt{+++-{}+} & 4.96 & 1.18 & 7.75 & 0.96 & 7.21 & 2.88 & 5.64 & 1.28 \\
\texttt{-{}+-{}-{}+} & 4.72 & 1.05 & 7.9 & 0.78 & 7.19 & 2.27 & 5.95 & 1.42 \\
\texttt{+++++} & 4.87 & 1.0 & 7.7 & 0.87 & 6.99 & 2.61 & 5.68 & 1.21 \\
\texttt{-{}+-{}++} & 4.78 & 1.03 & 7.78 & 0.87 & 7.33 & 2.56 & 5.87 & 1.41 \\
\texttt{-{}++-{}-{}} & 4.71 & 0.99 & 7.66 & 0.87 & 6.83 & 2.47 & 5.67 & 1.36 \\
\texttt{-{}-{}+++} & 4.69 & 1.02 & 7.65 & 0.88 & 7.0 & 2.47 & 5.81 & 1.47 \\
\texttt{-{}+-{}+-{}} & 4.7 & 1.18 & 7.6 & 1.01 & 7.02 & 2.76 & 5.67 & 1.49 \\
\texttt{++-{}-{}+} & 4.85 & 0.98 & 7.44 & 1.03 & 7.44 & 2.82 & 5.64 & 1.36 \\
\texttt{++-{}++} & 4.74 & 1.15 & 7.48 & 1.04 & 7.31 & 3.07 & 5.57 & 1.31 \\
\texttt{-{}-{}+-{}+} & 4.66 & 1.0 & 7.55 & 0.96 & 7.14 & 2.56 & 5.63 & 1.33 \\
\texttt{-{}+-{}-{}-{}} & 4.71 & 1.0 & 7.5 & 0.99 & 7.22 & 2.64 & 5.47 & 1.45 \\
\texttt{-{}+++-{}} & 4.55 & 0.93 & 7.56 & 0.9 & 7.05 & 2.58 & 5.7 & 1.47 \\
\texttt{+-{}+++} & 4.78 & 1.1 & 7.23 & 1.14 & 7.5 & 4.09 & 5.61 & 1.23 \\
\texttt{-{}-{}-{}-{}+} & 4.55 & 1.04 & 7.27 & 1.02 & 6.89 & 2.37 & 5.46 & 1.45 \\
\texttt{-{}-{}+-{}-{}} & 4.44 & 0.86 & 7.31 & 1.04 & 6.99 & 2.55 & 5.35 & 1.32 \\
\texttt{-{}-{}++-{}} & 4.5 & 1.02 & 7.24 & 1.2 & 6.66 & 2.86 & 5.34 & 1.37 \\
\texttt{-{}-{}-{}++} & 4.51 & 0.85 & 7.2 & 1.16 & 7.12 & 3.23 & 5.5 & 1.47 \\
\texttt{+++-{}-{}} & 4.57 & 0.93 & 6.99 & 1.31 & 7.46 & 2.89 & 5.47 & 1.49 \\
\texttt{++++-{}} & 4.56 & 1.01 & 6.94 & 1.23 & 7.11 & 2.27 & 5.26 & 1.36 \\
\texttt{+-{}+-{}+} & 4.55 & 0.98 & 6.57 & 1.44 & 7.68 & 2.91 & 5.43 & 1.14 \\
\texttt{-{}-{}-{}+-{}} & 4.45 & 0.93 & 6.58 & 1.21 & 7.08 & 3.14 & 5.21 & 1.26 \\
\texttt{-{}-{}-{}-{}-{}} & 4.27 & 1.02 & 6.7 & 1.4 & 7.47 & 3.95 & 5.09 & 1.2 \\
\texttt{++-{}-{}-{}} & 4.49 & 0.88 & 6.42 & 1.26 & 7.5 & 2.86 & 5.15 & 1.26 \\
\texttt{++-{}+-{}} & 4.39 & 0.85 & 6.44 & 1.21 & 7.53 & 2.51 & 5.42 & 1.29 \\
\texttt{+-{}-{}++} & 4.44 & 1.24 & 5.9 & 1.81 & 7.8 & 3.41 & 5.26 & 1.2 \\
\texttt{+-{}-{}-{}+} & 4.11 & 1.46 & 5.31 & 2.03 & 7.86 & 3.09 & 5.11 & 1.22 \\
\texttt{+-{}++-{}} & 3.83 & 1.18 & 4.85 & 1.72 & 7.62 & 3.12 & 4.86 & 1.13 \\
\texttt{+-{}+-{}-{}} & 3.49 & 1.32 & 4.72 & 2.06 & 7.9 & 2.53 & 4.54 & 1.23 \\
\texttt{+-{}-{}+-{}} & 3.47 & 1.2 & 4.06 & 1.88 & 8.11 & 3.03 & 4.76 & 1.15 \\
\texttt{+-{}-{}-{}-{}} & 3.14 & 1.48 & 3.75 & 2.07 & 7.74 & 2.9 & 4.54 & 1.21 \\
\hline
\end{tabular}

\caption{Creativity scores on SCIENTIFIC for agents with different Big Five personality traits using Llama3.1-8B. Scores include the four TTCT creativity assessment dimensions: ORIGINALITY, ELABORATION, FLUENCY, and FLEXIBILITY.}
\label{sin:SCI:llam}
\end{table*}

\begin{table*}
\centering

\begin{tabular}{lcccccccc}
\hline
\multirow{2}{*}{Trait} & \multicolumn{2}{c}{ORIGINALITY} & \multicolumn{2}{c}{ELABORATION} & \multicolumn{2}{c}{FLUENCY} & \multicolumn{2}{c}{FLEXIBILITY} \\
\cline{2-9}
 & Mean & Std. & Mean & Std. & Mean & Std. & Mean & Std. \\
\hline
\texttt{-{}+-{}++} & 3.75 & 0.79 & 7.74 & 1.01 & 6.92 & 1.26 & 6.76 & 1.19 \\
\texttt{+++++} & 3.74 & 0.72 & 7.59 & 0.9 & 6.79 & 0.9 & 6.61 & 0.95 \\
\texttt{-{}++++} & 3.61 & 0.76 & 7.69 & 0.9 & 7.25 & 1.05 & 7.03 & 1.08 \\
\texttt{-{}++-{}+} & 3.62 & 0.83 & 7.67 & 0.92 & 6.81 & 0.91 & 6.68 & 0.9 \\
\texttt{+++-{}+} & 3.69 & 0.63 & 7.5 & 0.87 & 6.44 & 0.86 & 6.18 & 0.9 \\
\texttt{-{}+-{}-{}+} & 3.46 & 0.73 & 7.53 & 1.11 & 6.63 & 1.07 & 6.35 & 0.96 \\
\texttt{++-{}-{}+} & 3.75 & 0.79 & 7.11 & 1.21 & 6.9 & 1.25 & 6.19 & 1.06 \\
\texttt{-{}-{}+-{}+} & 3.38 & 0.78 & 7.35 & 1.02 & 6.36 & 0.94 & 6.11 & 0.81 \\
\texttt{-{}+++-{}} & 3.46 & 0.73 & 7.21 & 0.94 & 6.6 & 1.17 & 6.48 & 1.15 \\
\texttt{-{}+-{}+-{}} & 3.69 & 0.87 & 6.88 & 1.29 & 5.89 & 1.03 & 5.75 & 0.9 \\
\texttt{++-{}++} & 3.56 & 0.75 & 6.96 & 1.19 & 6.65 & 1.39 & 6.08 & 1.25 \\
\texttt{-{}++-{}-{}} & 3.39 & 0.77 & 6.83 & 1.01 & 6.33 & 1.09 & 6.17 & 1.02 \\
\texttt{+++-{}-{}} & 3.6 & 0.81 & 6.44 & 1.26 & 6.57 & 1.34 & 5.96 & 1.07 \\
\texttt{++++-{}} & 3.56 & 0.77 & 6.46 & 1.2 & 6.65 & 1.37 & 6.04 & 0.97 \\
\texttt{-{}+-{}-{}-{}} & 3.24 & 0.74 & 6.36 & 0.91 & 6.14 & 1.12 & 5.71 & 0.91 \\
\texttt{-{}-{}+++} & 3.16 & 0.81 & 5.96 & 1.26 & 6.55 & 1.56 & 5.62 & 1.02 \\
\texttt{-{}-{}+-{}-{}} & 3.0 & 0.75 & 5.83 & 1.08 & 6.12 & 1.44 & 5.48 & 1.04 \\
\texttt{+-{}+++} & 3.6 & 0.72 & 5.16 & 0.97 & 6.77 & 1.7 & 5.49 & 1.21 \\
\texttt{-{}-{}-{}++} & 3.18 & 0.73 & 5.26 & 1.04 & 6.52 & 1.53 & 5.48 & 1.07 \\
\texttt{+-{}-{}++} & 3.46 & 0.89 & 4.93 & 0.97 & 6.71 & 1.68 & 5.31 & 0.88 \\
\texttt{+-{}+-{}+} & 3.47 & 0.74 & 4.87 & 0.98 & 6.6 & 1.9 & 5.15 & 1.08 \\
\texttt{-{}-{}++-{}} & 2.96 & 0.71 & 5.28 & 1.13 & 6.37 & 1.58 & 5.28 & 1.14 \\
\texttt{++-{}+-{}} & 3.33 & 0.8 & 4.83 & 0.78 & 6.62 & 1.59 & 5.34 & 0.95 \\
\texttt{++-{}-{}-{}} & 3.11 & 0.76 & 4.62 & 0.69 & 6.74 & 1.75 & 5.08 & 0.82 \\
\texttt{-{}-{}-{}-{}+} & 2.95 & 0.71 & 4.73 & 1.12 & 6.55 & 1.88 & 4.84 & 1.04 \\
\texttt{-{}-{}-{}+-{}} & 3.03 & 0.75 & 4.35 & 0.68 & 6.36 & 1.65 & 5.01 & 0.96 \\
\texttt{+-{}-{}-{}+} & 3.02 & 0.88 & 4.06 & 1.08 & 7.32 & 1.98 & 4.86 & 0.87 \\
\texttt{+-{}++-{}} & 2.97 & 0.83 & 3.83 & 0.85 & 7.32 & 1.97 & 4.6 & 1.06 \\
\texttt{+-{}-{}+-{}} & 2.81 & 0.77 & 3.45 & 0.9 & 7.77 & 1.8 & 4.46 & 0.95 \\
\texttt{+-{}+-{}-{}} & 2.68 & 0.79 & 3.39 & 0.86 & 8.0 & 1.81 & 4.04 & 1.04 \\
\texttt{-{}-{}-{}-{}-{}} & 2.34 & 0.55 & 3.21 & 0.86 & 8.4 & 1.79 & 3.76 & 0.85 \\
\texttt{+-{}-{}-{}-{}} & 2.45 & 0.67 & 2.73 & 0.8 & 8.13 & 1.38 & 3.72 & 0.81 \\
\hline
\end{tabular}

\caption{Creativity scores on the SIMILARITIES fo for agents with different Big Five personality traits using Qwen2.5-32B. Scores include the four TTCT creativity assessment dimensions: ORIGINALITY, ELABORATION, FLUENCY, and FLEXIBILITY.}
\label{sin:sim:32B}
\end{table*}

\begin{table*}
\centering

\begin{tabular}{lcccccccc}
\hline
\multirow{2}{*}{Trait} & \multicolumn{2}{c}{ORIGINALITY} & \multicolumn{2}{c}{ELABORATION} & \multicolumn{2}{c}{FLUENCY} & \multicolumn{2}{c}{FLEXIBILITY} \\
\cline{2-9}
 & Mean & Std. & Mean & Std. & Mean & Std. & Mean & Std. \\
\hline
\texttt{-{}++++} & 3.27 & 0.8 & 7.52 & 0.95 & 7.16 & 1.32 & 6.98 & 1.3 \\
\texttt{+++-{}+} & 3.36 & 0.79 & 7.37 & 0.96 & 6.3 & 0.97 & 6.26 & 0.99 \\
\texttt{-{}++-{}+} & 3.39 & 0.77 & 7.27 & 0.95 & 6.96 & 1.11 & 6.78 & 1.11 \\
\texttt{+++++} & 3.29 & 0.77 & 7.27 & 1.02 & 6.41 & 0.85 & 6.37 & 0.87 \\
\texttt{-{}+-{}-{}+} & 3.23 & 0.76 & 7.26 & 1.0 & 6.93 & 1.22 & 6.77 & 1.14 \\
\texttt{-{}-{}+-{}+} & 3.42 & 0.76 & 7.01 & 1.05 & 6.74 & 1.12 & 6.7 & 1.11 \\
\texttt{-{}+-{}++} & 3.28 & 0.78 & 6.95 & 0.98 & 6.76 & 0.97 & 6.56 & 0.93 \\
\texttt{++-{}++} & 3.52 & 0.92 & 6.7 & 1.2 & 6.3 & 1.08 & 6.16 & 1.1 \\
\texttt{-{}+-{}-{}-{}} & 3.32 & 0.73 & 6.88 & 1.01 & 6.3 & 0.9 & 6.25 & 0.91 \\
\texttt{-{}++-{}-{}} & 3.22 & 0.74 & 6.9 & 1.07 & 6.48 & 1.03 & 6.33 & 0.99 \\
\texttt{-{}+-{}+-{}} & 3.26 & 0.77 & 6.79 & 1.06 & 6.4 & 1.04 & 6.32 & 1.07 \\
\texttt{-{}+++-{}} & 3.31 & 0.73 & 6.65 & 1.06 & 6.4 & 0.82 & 6.34 & 0.83 \\
\texttt{++-{}-{}+} & 3.39 & 0.93 & 6.56 & 1.05 & 5.87 & 1.1 & 5.65 & 1.04 \\
\texttt{-{}-{}-{}-{}+} & 3.32 & 0.77 & 6.56 & 0.97 & 6.47 & 1.15 & 6.27 & 1.05 \\
\texttt{-{}-{}+++} & 3.11 & 0.73 & 6.72 & 0.95 & 6.73 & 1.16 & 6.56 & 1.09 \\
\texttt{-{}-{}++-{}} & 3.4 & 0.71 & 6.37 & 0.87 & 6.2 & 1.0 & 6.15 & 0.99 \\
\texttt{-{}-{}+-{}-{}} & 3.2 & 0.81 & 6.51 & 0.96 & 6.37 & 1.11 & 6.26 & 1.05 \\
\texttt{-{}-{}-{}++} & 3.3 & 0.75 & 6.4 & 0.92 & 6.47 & 1.11 & 6.38 & 1.07 \\
\texttt{-{}-{}-{}-{}-{}} & 3.34 & 0.76 & 6.22 & 0.87 & 6.06 & 0.98 & 6.01 & 0.99 \\
\texttt{+++-{}-{}} & 3.2 & 0.77 & 6.35 & 1.13 & 5.8 & 1.1 & 5.65 & 1.02 \\
\texttt{-{}-{}-{}+-{}} & 3.21 & 0.74 & 6.28 & 0.91 & 6.32 & 1.07 & 6.18 & 1.03 \\
\texttt{++++-{}} & 3.33 & 0.85 & 6.03 & 0.91 & 6.07 & 1.07 & 5.78 & 0.97 \\
\texttt{++-{}+-{}} & 3.38 & 1.08 & 5.74 & 0.92 & 6.0 & 1.22 & 5.51 & 0.88 \\
\texttt{++-{}-{}-{}} & 3.19 & 0.78 & 5.51 & 0.99 & 6.01 & 1.49 & 5.22 & 1.15 \\
\texttt{+-{}+++} & 3.03 & 0.71 & 5.54 & 1.17 & 6.39 & 1.35 & 5.77 & 1.01 \\
\texttt{+-{}-{}++} & 3.07 & 0.72 & 4.59 & 0.76 & 6.54 & 1.55 & 5.52 & 1.07 \\
\texttt{+-{}+-{}+} & 3.05 & 0.75 & 4.46 & 0.9 & 7.25 & 2.13 & 4.93 & 0.94 \\
\texttt{+-{}-{}-{}+} & 2.73 & 0.71 & 3.99 & 0.93 & 7.51 & 2.18 & 4.54 & 1.02 \\
\texttt{+-{}++-{}} & 2.72 & 0.86 & 3.81 & 0.72 & 7.25 & 1.88 & 4.67 & 1.1 \\
\texttt{+-{}-{}+-{}} & 2.5 & 0.75 & 3.39 & 0.75 & 8.07 & 1.97 & 4.24 & 0.94 \\
\texttt{+-{}+-{}-{}} & 2.37 & 0.6 & 3.29 & 0.84 & 7.7 & 2.14 & 4.01 & 0.88 \\
\texttt{+-{}-{}-{}-{}} & 2.35 & 0.59 & 3.26 & 0.76 & 7.75 & 2.05 & 4.12 & 0.93 \\
\hline
\end{tabular}

\caption{Creativity scores on the SIMILARITIES fo for agents with different Big Five personality traits using Qwen2.5-14B. Scores include the four TTCT creativity assessment dimensions: ORIGINALITY, ELABORATION, FLUENCY, and FLEXIBILITY.}
\label{sin:sim:14B}
\end{table*}

\begin{table*}
\centering

\begin{tabular}{lcccccccc}
\hline
\multirow{2}{*}{Trait} & \multicolumn{2}{c}{ORIGINALITY} & \multicolumn{2}{c}{ELABORATION} & \multicolumn{2}{c}{FLUENCY} & \multicolumn{2}{c}{FLEXIBILITY} \\
\cline{2-9}
 & Mean & Std. & Mean & Std. & Mean & Std. & Mean & Std. \\
\hline
\texttt{-{}++++} & 3.33 & 0.76 & 7.45 & 0.9 & 8.61 & 1.69 & 7.41 & 1.81 \\
\texttt{-{}++-{}+} & 3.59 & 0.92 & 6.94 & 0.97 & 7.36 & 1.46 & 6.39 & 1.43 \\
\texttt{-{}+-{}-{}+} & 3.84 & 0.83 & 6.51 & 1.16 & 6.92 & 1.59 & 5.82 & 1.19 \\
\texttt{-{}+-{}++} & 3.47 & 0.76 & 6.3 & 1.01 & 6.96 & 1.54 & 6.26 & 1.49 \\
\texttt{-{}+++-{}} & 2.92 & 0.7 & 6.08 & 0.84 & 6.81 & 1.94 & 5.9 & 1.49 \\
\texttt{-{}+-{}+-{}} & 3.27 & 0.8 & 5.7 & 0.78 & 6.8 & 1.72 & 5.77 & 1.3 \\
\texttt{-{}++-{}-{}} & 2.7 & 0.73 & 4.86 & 0.88 & 7.53 & 1.71 & 5.41 & 1.32 \\
\texttt{+++-{}+} & 3.14 & 0.97 & 4.24 & 0.9 & 6.89 & 1.61 & 5.07 & 1.2 \\
\texttt{+++++} & 3.08 & 0.87 & 4.16 & 0.78 & 7.09 & 1.65 & 5.41 & 1.27 \\
\texttt{++-{}-{}+} & 2.81 & 0.84 & 4.19 & 0.86 & 7.06 & 1.72 & 5.21 & 1.22 \\
\texttt{-{}-{}+++} & 2.68 & 0.72 & 4.29 & 0.91 & 7.69 & 1.47 & 5.7 & 1.26 \\
\texttt{++-{}++} & 3.03 & 0.85 & 3.94 & 0.89 & 7.32 & 1.48 & 5.52 & 1.32 \\
\texttt{-{}-{}+-{}+} & 2.62 & 0.64 & 4.34 & 0.87 & 7.93 & 1.72 & 5.45 & 1.24 \\
\texttt{-{}-{}-{}-{}+} & 2.65 & 0.67 & 4.03 & 0.83 & 8.14 & 1.84 & 5.66 & 1.52 \\
\texttt{+++-{}-{}} & 2.73 & 0.81 & 3.89 & 0.9 & 7.34 & 1.44 & 4.91 & 1.11 \\
\texttt{++++-{}} & 2.71 & 0.77 & 3.79 & 0.88 & 7.56 & 1.51 & 5.04 & 0.97 \\
\texttt{-{}-{}+-{}-{}} & 2.46 & 0.56 & 3.95 & 0.77 & 8.02 & 1.71 & 5.16 & 1.51 \\
\texttt{-{}+-{}-{}-{}} & 2.55 & 0.7 & 3.77 & 0.88 & 7.93 & 3.32 & 5.12 & 1.37 \\
\texttt{-{}-{}-{}++} & 2.5 & 0.62 & 3.8 & 0.82 & 7.59 & 1.67 & 5.5 & 1.31 \\
\texttt{+-{}+++} & 2.68 & 0.75 & 3.57 & 0.74 & 7.36 & 1.33 & 5.16 & 1.0 \\
\texttt{+-{}-{}++} & 2.68 & 0.69 & 3.56 & 0.71 & 7.5 & 1.52 & 5.12 & 1.18 \\
\texttt{-{}-{}++-{}} & 2.47 & 0.66 & 3.7 & 0.84 & 8.04 & 1.54 & 5.46 & 1.35 \\
\texttt{-{}-{}-{}-{}-{}} & 2.42 & 0.59 & 3.73 & 0.8 & 8.29 & 2.11 & 5.46 & 1.19 \\
\texttt{++-{}+-{}} & 2.52 & 0.67 & 3.55 & 0.92 & 7.65 & 1.57 & 5.02 & 1.09 \\
\texttt{+-{}-{}-{}+} & 2.42 & 0.72 & 3.38 & 0.82 & 7.61 & 1.41 & 5.09 & 1.16 \\
\texttt{+-{}+-{}+} & 2.47 & 0.62 & 3.29 & 0.71 & 7.48 & 1.67 & 5.22 & 1.12 \\
\texttt{++-{}-{}-{}} & 2.47 & 0.69 & 3.25 & 0.93 & 7.8 & 1.52 & 4.66 & 0.96 \\
\texttt{+-{}++-{}} & 2.47 & 0.61 & 3.24 & 0.67 & 7.39 & 1.64 & 4.93 & 1.18 \\
\texttt{+-{}+-{}-{}} & 2.45 & 0.61 & 3.26 & 0.81 & 7.65 & 1.44 & 4.71 & 1.05 \\
\texttt{-{}-{}-{}+-{}} & 2.34 & 0.55 & 3.19 & 0.81 & 8.56 & 2.03 & 4.82 & 1.32 \\
\texttt{+-{}-{}+-{}} & 2.42 & 0.62 & 3.08 & 0.76 & 7.6 & 1.73 & 4.92 & 1.15 \\
\texttt{+-{}-{}-{}-{}} & 2.3 & 0.63 & 3.08 & 0.73 & 8.63 & 1.94 & 4.89 & 1.11 \\
\hline
\end{tabular}

\caption{Creativity scores on the SIMILARITIES fo for agents with different Big Five personality traits using Llama3.1-8B. Scores include the four TTCT creativity assessment dimensions: ORIGINALITY, ELABORATION, FLUENCY, and FLEXIBILITY.}
\label{sin:SIMILARITIES:llama}
\end{table*}

\end{document}